\documentclass[journal]{IEEEtran}

\usepackage{cite}
\usepackage{amsmath,amssymb,amsfonts}
\usepackage{graphicx}
\usepackage{booktabs}
\usepackage{multirow}
\usepackage{array}
\usepackage{caption}
\usepackage[caption=false,font=footnotesize,labelfont=bf]{subfig}
\usepackage{xcolor}
\usepackage[hidelinks]{hyperref}
\usepackage{url}
\usepackage{tabularx}

\usepackage{algorithm}
\usepackage{algpseudocode}
\usepackage{physics}

\graphicspath{{figures_limits/}{catalogue_figures/}{./}}
\newcommand{\QieoFixK}{174}
\newcommand{\QieoFixN}{256}
\newcommand{\QieoFixPct}{68.0}
\newcommand{\QieoFixLo}{62.0}
\newcommand{\QieoFixHi}{73.4}
\newcommand{\QieoAdpK}{166}
\newcommand{\QieoAdpN}{256}
\newcommand{\QieoAdpPct}{64.8}
\newcommand{\QieoAdpLo}{58.8}
\newcommand{\QieoAdpHi}{70.4}
\newcommand{\QieoRealK}{194}
\newcommand{\QieoRealN}{256}
\newcommand{\QieoRealPct}{75.8}
\newcommand{\QieoRealLo}{70.2}
\newcommand{\QieoRealHi}{80.6}
\newcommand{\QieoBestK}{203}
\newcommand{\QieoBestN}{256}
\newcommand{\QieoBestPct}{79.3}
\newcommand{\QieoBestLo}{73.9}
\newcommand{\QieoBestHi}{83.8}
\newcommand{\GaRealK}{111}
\newcommand{\GaRealN}{256}
\newcommand{\GaRealPct}{43.4}
\newcommand{\GaRealLo}{37.4}
\newcommand{\GaRealHi}{49.5}
\newcommand{\GaBinK}{105}
\newcommand{\GaBinN}{256}
\newcommand{\GaBinPct}{41.0}
\newcommand{\GaBinLo}{35.2}
\newcommand{\GaBinHi}{47.1}
\newcommand{\GaBestK}{119}
\newcommand{\GaBestN}{256}
\newcommand{\GaBestPct}{46.5}
\newcommand{\GaBestLo}{40.5}
\newcommand{\GaBestHi}{52.6}
\newcommand{\CmaK}{159}
\newcommand{\CmaN}{256}
\newcommand{\CmaPct}{62.1}
\newcommand{\CmaLo}{56.0}
\newcommand{\CmaHi}{67.8}

\newcommand{\nQfail}{62}
\newcommand{\nGaRescue}{1}
\newcommand{\nCmaRescue}{14}

\newcommand{\HardQieoN}{63}
\newcommand{\HardQieoPct}{57.1}

\newcommand{\HardCmaPct}{58.7}

\newcommand{\cmaMedEv}{670}
\newcommand{\cmaMaxEv}{7849}
\newcommand{\gaMedEv}{88500}
\newcommand{\cmaCapFrac}{0.0134}
\newcommand{\cmaPerM}{583.6}
\newcommand{\gaPerM}{4.725}

\newcommand{\nCore}{44}
\newcommand{\nAllSolved}{88}
\newcommand{\coreMedD}{2}

\newcommand{\nOnlyQieo}{49}
\newcommand{\nOnlyCma}{12}
\newcommand{\nBothQC}{144}
\newcommand{\nVbsQC}{205}
\newcommand{\nVbsAll}{212}
\newcommand{\gridOn}{91}
\newcommand{\gridChecked}{254}
\newcommand{\qieoMedEv}{87003}
\newcommand{\qieoMeanEv}{427347}

\newcommand{\qieoAtCap}{16}
\newcommand{\qieoTotMEv}{109}
\newcommand{\gaTotMEv}{25}
\newcommand{\qieoMedGen}{87}
\newcommand{\qieoCapFrac}{1.7}
\newcommand{\qieoPerM}{1.79}

\newcommand{\cmaOverQieo}{326}

\newcommand{\bandQEight}{46.5}
\newcommand{\bandCEight}{51.2}
\newcommand{\bandGEight}{6.2}
\newcommand{\bandQTwelve}{10.9}
\newcommand{\bandCTwelve}{43.4}

\newcommand{\flipExp}{-7.8}
\newcommand{\ExcQieoReal}{85}

\newcommand{\ExcGaBest}{28}
\newcommand{\ExcCma}{90}

\begin{document}

\title{Landscape Limits of\\
Quantum-Inspired Evolutionary Optimization\\
across 256 continuous functions}

\author{Rishi~Govind$^{*}$,
		Ferdin~Sagai~Don~Bosco$^{*}$,
        Kasturi~Venkata~Srikanth,
        Aman~Mittal,
        Abhishek~Singh,
        Aditya~Singh
        and~Abhishek Chopra
\thanks{$^{*}$These authors contributed equally to this research.}%
\thanks{All authors are with BosonQ Psi Corporation, Bangalore, Karnataka, India.}
\thanks{
Rishi~Govind(ORCID: 0009-0002-1588-2977, rishigovind36@gmail.com);
F.~S.~D. Bosco (ORCID: 0000-0002-5507-5304, ferdindon@bqpsim.com);
K.~V. Srikanth (ORCID: 0000-0002-0159-773X, kasturi.srikanth@bqpsim.com);
A.~Mittal (ORCID: 0009-0002-7737-6893, aman.mittal@bqpsim.com);
Abhishek Singh (ORCID: 0009-0009-7140-7047, abhishek.singh@bqpsim.com);
Aditya Singh (ORCID: 1234-5678-9012, singh.aditya@bqpsim.com);
A.~Chopra (ORCID: 0000-0002-1196-7765, abhishekchopra@bqpsim.com).}}

\markboth{}%
{Bosco \MakeLowercase{\textit{et al.}}: Coverage or Curvature?}

\maketitle

\begin{abstract}
Quantum-inspired evolutionary optimization (QIEO) represents design variables as a set of qubits and searches a continuous, multi-dimensional landscape through rotation of the qubit's amplitude pair. Every generation rotates those amplitudes toward a single elite, which corresponds to that generation's best. The update is cheap, almost parameter-free, and well-suited for massive parallel implementation, which has encouraged its adoption in engineering, design, and planning applications. However, there are critical issues with this formulation, principally, the treatment of design variables as independent probability components which make it incapable of exploiting local curvature, anisotropy, or variable coupling. Despite this, QIEO is believed to hold promise, and has been used extensively to solve real-world problems, with significant qualitative and computational advantage over its classical counterpart, Genetic Algorithm (GA). 

To facilitate widespread adoption and bolster confidence, characterization of QIEO's limits is essential and is the principal motivation of this research. 

A collection of 256 (actually 508; 256 unshifted + 252 shifted, 4 could not be shifted) continuous function are selected from the prior works, in such a way that they represent eleven landscape characteristics, namely continuity, differentiability, separability, scalability, modality, convexity, conditioning, symmetry, maximum dimensionality, dimension dependency, and the coupling pattern of the design variables. These functions are then solved by three QIEO variants, two GA encodings and Hansen's Covariance Matrix Adaptation Evolution Strategy (CMA-ES).

The results are evaluated in terms of computational cost, solution precision, and specialization across landscape characteristics. They identify the conditions under which QIEO provides competitive performance, clarify where its independent-variable representation becomes limiting, and establish whether particular QIEO variants offer advantages for specific landscape characteristics.

The coded catalogue and per-function tables are reprinted in the Appendix. A package is also provided that enables readers to replicate this study to ensure transparency.
\end{abstract}


\section{Introduction}

\IEEEPARstart{Q}{uantum}-Inspired Evolutionary Optimization (QIEO) has emerged as a powerful meta-heuristic for complex optimization problems arising in engineering, design, and logistics\cite{eswara2024qieo, mittal__aman_ac24343c}. The key appeal is attributed to QIEO's unique independent evolutionary strategy and reliance on a single parameter/parameter-free rotation operator, which effectively mitigates common limitations of conventional meta-heuristics, particularly in terms of consistency and convergence rate  \cite{eswara2024qieo}.

The rotation operator acting on a single pair of amplitudes is about as small as a search operator can be~\cite{narayanan1996,han2002qea}. QIEO takes that operator literally and applies it to a real multi-dimensional search space. Every coordinate (often referred to as gene) of every individual (often referred to as chromosome) carries a pair \((\alpha,\beta)\) with \(|\alpha|^2+|\beta|^2=1\); a classical design vector is observed from those amplitudes; the generation's best chromosome is identified; and each amplitude pair is rotated by an angle \(\Delta\theta\) in the direction that makes the elite's bit or value more likely next time~\cite{eswara2024qieo}. 

There is almost no hyperparameter surface to tune, and rotating a population of one thousand individuals is arithmetically trivial and amenable to massive parallelism. This is naturally alluring to engineering and logistic challenges, and consequently, the method has been applied to real-world challenges including, cargo loading, supply chain, route planning, material design, satellite constellation placement, and structural optimization.

However, QIEO is not without its flaws. A locally applied rotation cannot represent a covariance, which limits the search to a single impetus. This manifests as a pull that draws the whole population toward one elite. QIEO cannot keep two competing basins alive, so it cannot hedge. Genetic algorithms made a structurally similar sacrifice with crossover and mutation and were criticised for exactly these reasons~\cite{holland1975,goldberg1989ga,salomon1996}. Covariance Matrix Adaptation Evolution Strategy (CMA-ES) takes the opposite approach, adapting a full covariance matrix from the history of successful search steps to capture second-order structure in the objective landscape~\cite{hansen2001cma,hansen2003reducing,hansen2016tutorial}.

The central research question is: \textbf{Under what conditions does QIEO fail to achieve competitive optimization performance?}

To answer this, the research considers 256 continuous function consisting of the 175 functions surveyed by Jamil and Yang~\cite{jamil2013survey} plus 81 additions from Gavana's collection~\cite{gavana2013}. Each of these function are categorised by eleven characteristics: continuity, differentiability, separability, scalability, modality, convexity, conditioning, symmetry, maximum dimensionality, whether \(f^\star\) depends on the dimension, and the coupling pattern of the design variables. This suite is large enough to stratify and honest enough to be labelled independently of the optimizer traces.

A dedicated campaign is performed in which 3 variants of QIEO, 2 encodings of GA, and CMA-ES algorithms are used to solve all 256 unshifted and 252 shifted functions. The key findings are summarized as follows,

Cost-wise, CMA-ES reaches \(\CmaPct\%\) strict success at a median of \(\cmaMedEv\) evaluations per function, \(\cmaCapFrac\%\) of the cap, a yield of \(\cmaPerM\) solved functions per million evaluations against \(\qieoPerM\) for QIEO, which spends a median of \(\qieoMedEv\).

Second, precision: the two solvers fail differently, and the ranking \emph{reverses} at a tolerance of \(10^{\flipExp}\). At \(10^{-8}\), CMA-ES leads \(\bandCEight\%\) to \(\bandQEight\%\).

Third, specialization: QIEO's \(\nOnlyQieo\) exclusive wins are almost all two-dimensional, densely multimodal landscapes solved by population coverage, while all \(\nOnlyCma\) exclusive CMA-ES wins are multimodal and mostly ill-conditioned, including nonlinear-regression residuals where a learned metric is decisive. A multivariate model shows the eleven hand-assigned labels collapse to two effective axes, and that the two solvers split along them: QIEO degrades with dimension and is indifferent to conditioning; CMA-ES is indifferent to dimension and degrades with multimodality. Only modality and convexity survive multiple-comparison correction as univariate predictors, so three of the five ``significant'' properties in a naive analysis are proxies. We also identify \(\nCore\) functions no solver reaches, a family signature rather than a difficulty scale, and six functions whose published \(f^\star\) is undercut by nearly every solver and is therefore probably wrong.

Our contributions are as follows.
\begin{enumerate}
\item \textbf{A labelled map of when QIEO succeeds.} We report strict success for three QIEO variants, a matched GA, and default CMA-ES on 256 continuous functions, unshifted and shifted, stratified by eleven landscape characteristics.
\item \textbf{A cost-normalized comparison.} QIEO's evaluation counts are measured, not assumed. Its higher hit rate costs two to three orders of magnitude more sampling than CMA-ES; both GA encodings are worse on cost and on quality at once.
\item \textbf{Accuracy profiles, not a single gate.} Which solver ``wins'' depends on the tolerance. The ranking reverses at a residual of \(10^{\flipExp}\), and the profiles show why.
\item \textbf{Which landscape labels actually matter.} Fitted jointly, the eleven expert tags collapse to two axes. QIEO weakens with dimension and is indifferent to conditioning; CMA-ES is the other way around.
\item \textbf{What remains unsolved, and which references are wrong.} \(\nCore\) names are missed by every solver, grouped by problem family rather than by difficulty. Six published \(f^\star\) values are undercut by nearly every method and are probably incorrect.
\item \textbf{An archival catalogue.} The coded objectives and per-configuration campaign tables are reprinted in the Appendix, and a Docker package is provided so the runs can be repeated.
\end{enumerate}

\section{Related Work}

Han and Kim introduced the qubit chromosome and its lookup-table rotation for combinatorial problems, establishing the template that continuous QIEO later adopted~\cite{han2002qea}; Narayanan and Moore had earlier proposed quantum-inspired genetic operators~\cite{narayanan1996}. Continuous variants port the rotation onto a real box and have been reported to reduce wall-clock time relative to a GA on small collections of GPU-resident test functions~\cite{eswara2024qieo}. Those studies establish feasibility. None is a landscape survey, and none reports budget-normalized comparisons, which is the gap this paper addresses.

CMA-ES occupies the opposite design point. It is a derandomized evolution strategy that estimates a covariance matrix and a global step size from the ranks of recent successful steps~\cite{hansen2001cma,hansen2003reducing}. On quadratic bowls and narrow ridges it is very hard to beat, and its behaviour under ill-conditioning is the property most often cited in its favour~\cite{hansen2016tutorial}. Its known liability is a small default population on strongly multimodal landscapes, which is why restart schemes with growing populations exist~\cite{auger2005performance,hansen2010comparing}. 

Genetic algorithms remain the ancestral population search in both real and binary codings~\cite{holland1975,goldberg1989ga,deb2001}. Salomon's demonstration that rotating a separable benchmark is sufficient to break many GAs is the direct precedent for asking whether a coordinate-wise quantum gate inherits the same weakness~\cite{salomon1996}. Our gate-sensitivity and budget analyses are in that tradition.


On the benchmark side, Jamil and Yang catalogued 175 continuous functions with analytic minima and a coarse taxonomy~\cite{jamil2013survey}, and Gavana's collection supplies further continuous functions widely used in global-optimization testing~\cite{gavana2013}. The CEC and BBOB families instead emphasize shifted, rotated and hybridized compositions~\cite{liang2013cec,hansen2010comparing}. We keep the continuous functions precisely because a failure can then be attributed to a documented property rather than to a random orthogonal mixing, at the acknowledged cost of a suite whose dimension distribution is low and whose composition is historically, not statistically, determined. Exploratory landscape analysis is the methodological parent of the stratification used here~\cite{mersmann2011ela}, and we adopt its spirit while using published labels instead of sampled features.

\section{Methods}

\subsection{Quantum-inspired evolutionary optimization} \label{sec:qieo}

Each of the \(D\) design variables of each individual carries a qubit amplitude pair \((\alpha,\beta)\) with \(|\alpha|^2+|\beta|^2=1\), so a population of \(N_p\) individuals is a \(N_p\times D\) array of such pairs\footnote{Note: If the design variables are real and the encoding is binary, each design variable is converted to a binary string and each binary element is represented as a qubit. Thus, the size would be \(N_p\times D\times s\) array, where s is the encoding length.}

Quantum-Inspired Evolutionary Optimization (QIEO) is a population-based, stochastic optimization method that uses quantum-mechanical concepts as a classical probabilistic search representation. It does not require quantum hardware: quantum states, measurement, and quantum gates are emulated on conventional computing platforms. In contrast to a conventional genetic algorithm, QIEO evolves probability amplitudes rather than directly evolving a population of fixed classical chromosomes.

Let the constrained optimization problem be written as
\begin{equation}
    \min_{\mathbf{x} \in \Omega} f(\mathbf{x}),
    \qquad
    \Omega
    =
    \prod_{j=1}^{D}[l_j,u_j],
    \label{eq:qieo_problem}
\end{equation}
where \(f(\mathbf{x})\) is the objective function, \(D\) is the number of decision variables, and \(l_j\) and \(u_j\) denote the lower and upper bounds of the \(j\)-th variable, respectively. QIEO maintains a quantum-inspired population whose individuals encode probability distributions over classical candidate solutions. A classical population is generated through repeated observation of this underlying quantum-inspired population.

The fundamental unit of representation is a quantum-inspired bit, or qubit, expressed as
\begin{equation}
    \ket{q}
    =
    \alpha \ket{0}
    +
    \beta \ket{1}
    =
    \begin{bmatrix}
        \alpha \\
        \beta
    \end{bmatrix},
    \label{eq:qieo_qubit_state}
\end{equation}
where \(\alpha\) and \(\beta\) are probability amplitudes satisfying the normalization constraint
\begin{equation}
    |\alpha|^2 + |\beta|^2 = 1.
    \label{eq:qieo_normalization}
\end{equation}
Upon observation, the qubit produces a classical binary value according to
\begin{equation}
    b =
    \begin{cases}
        0, & \text{with probability } |\alpha|^2, \\
        1, & \text{with probability } |\beta|^2.
    \end{cases}
    \label{eq:qieo_measurement}
\end{equation}

An individual consisting of \(m\) qubits is represented as
\begin{equation}
    \mathbf{q}_j
    =
    \left[
        \begin{array}{cccc}
            \alpha_{j,1} & \alpha_{j,2} & \cdots & \alpha_{j,m} \\
            \beta_{j,1}  & \beta_{j,2}  & \cdots & \beta_{j,m}
        \end{array}
    \right],
    \qquad
    j=1,\ldots,N_{\mathrm{pop}},
    \label{eq:qieo_individual}
\end{equation}
where \(N_{\mathrm{pop}}\) denotes the number of quantum-inspired individuals. A complete quantum-inspired population is therefore
\begin{equation}
    \mathcal{Q}^{(g)}
    =
    \left\{
        \mathbf{q}_1^{(g)},
        \mathbf{q}_2^{(g)},
        \ldots,
        \mathbf{q}_{N_{\mathrm{pop}}}^{(g)}
    \right\}.
    \label{eq:qieo_population}
\end{equation}

The initial state is typically selected as a uniform superposition,
\begin{equation}
    \alpha_{j,i}^{(0)}
    =
    \beta_{j,i}^{(0)}
    =
    \frac{1}{\sqrt{2}},
    \label{eq:qieo_initialization}
\end{equation}
for all individuals \(j\) and qubit positions \(i\). This initialization gives equal probability of observing zero or one at every qubit position and therefore produces broad initial sampling of the encoded search space.

Two representations are considered: binary-coded QIEO and real-coded QIEO. In the binary formulation, each design variable is represented by a fixed-length binary substring. For a \(D\)-dimensional optimization problem with \(B\) bits assigned to each variable, the chromosome length is
\begin{equation}
    m = BD.
    \label{eq:qieo_binary_chromosome_length}
\end{equation}

For the present formulation, \(B=16\) bits are used for each decision variable. Thus, an observed binary chromosome contains \(16D\) bits. 

A QIEO generation proceeds in three stages. First, a classical population is \emph{observed} from the quantum-inspired population. Under the binary encoding, each qubit collapses to a bit according to Eq.~\eqref{eq:qieo_measurement}; the resulting binary chromosome is subsequently decoded into a design vector using Eq.~\eqref{eq:qieo_binary_decoding}.

Second, the observed candidates are evaluated using the objective function. Let
\begin{equation}
    \mathcal{P}^{(g)}
    =
    \left\{
        \mathbf{x}_1^{(g)},
        \mathbf{x}_2^{(g)},
        \ldots,
        \mathbf{x}_{N_{\mathrm{pop}}}^{(g)}
    \right\}
    \label{eq:qieo_classical_population}
\end{equation}
denote the observed classical population at generation \(g\). For a minimization problem, the generation-best chromosome is
\begin{equation}
    \mathbf{b}^{(g)}
    =
    \arg\min_{\mathbf{x}\in\mathcal{P}^{(g)}}
    f(\mathbf{x}).
    \label{eq:qieo_generation_elite}
\end{equation}

The elite solution is retained as the reference chromosome for the subsequent amplitude update. The use of a persistent elite ensures that search updates are directed toward the best candidate identified during the optimization process.

Third, every qubit amplitude pair is rotated toward the elite chromosome. For each qubit, the update is expressed using a two-dimensional rotation matrix:
\begin{equation}
\begin{bmatrix}
    \alpha' \\
    \beta'
\end{bmatrix}
=
\begin{bmatrix}
    \cos (\theta + \Delta\theta) & -\sin(\theta + \Delta\theta) \\
    \sin (\theta + \Delta\theta) & \phantom{-}\cos(\theta + \Delta\theta)
\end{bmatrix}
\begin{bmatrix}
    \alpha \\
    \beta
\end{bmatrix}.
\label{eq:rot}
\end{equation}
The change in rotation angle \(\Delta\theta\) controls the magnitude of the probability-amplitude update. Its sign is selected such that the probability of observing the elite value at the corresponding coordinate increases after rotation.

For binary QIEO, if the elite bit at qubit position \(i\) is \(b_i^{\star}\), the update direction is selected to increase
\begin{equation}
    \Pr(b_i=b_i^{\star})
    =
    \begin{cases}
        |\alpha_i|^2, & b_i^{\star}=0, \\
        |\beta_i|^2, & b_i^{\star}=1.
    \end{cases}
    \label{eq:qieo_elite_probability}
\end{equation}
The sign of the rotation is therefore determined from the observed bit and the corresponding bit of the elite chromosome. If the observed and elite values agree, no corrective directional change is required; if they differ, the rotation is chosen to increase the likelihood of generating the elite bit in future observations.

The rotation operator preserves normalization because the rotation matrix is orthogonal:
\begin{equation}
    \left(\alpha'\right)^2
    +
    \left(\beta'\right)^2
    =
    \alpha^2+\beta^2
    =
    1.
    \label{eq:qieo_rotation_normalization}
\end{equation}
Hence, the qubit remains a valid probabilistic representation after every update.

Equation~\eqref{eq:rot} is the entire search operator. Unlike conventional evolutionary algorithms, QIEO does not employ explicit crossover or mutation operators. Instead, exploration arises from probabilistic observation of incompletely converged amplitude states, whereas exploitation is introduced through repeated small rotations toward the elite chromosome. A sufficiently small rotation angle prevents immediate collapse of the population onto the elite and retains a nonzero probability of observing alternatives in the search domain.

The QIEO update acts independently on each coordinate or bit position. For a binary chromosome, the implied sampling distribution is factored as
\begin{equation}
    p(\mathbf{b})
    =
    \prod_{i=1}^{m}
    p_i(b_i),
    \label{eq:qieo_factorized_distribution}
\end{equation}
where \(p_i(b_i)\) is determined only by the amplitude pair \((\alpha_i,\beta_i)\).

This factorized representation has an important structural consequence: standard QIEO cannot explicitly represent correlations, linkage, or covariance between distinct coordinates. In particular, it cannot directly learn coupled search directions of the form
\begin{equation}
    x_i \uparrow
    \quad \Longleftrightarrow \quad
    x_j \downarrow,
\end{equation}
which are common in nonseparable optimization landscapes. Such dependencies can only be reflected indirectly through the currently selected elite, rather than through a joint probability model over multiple variables.

A second consequence follows from directing every amplitude update toward a single elite. The standard method does not preserve separate probabilistic modes for multiple high-quality basins. When two or more competing regions contain promising candidates, their information is compressed into one elite reference solution at each update. Consequently, QIEO may be effective for broad coordinate-wise exploration but can be structurally limited on strongly nonseparable objectives or multimodal landscapes that require maintaining several distinct basins simultaneously.

\begin{figure}[!t]
\centering
\includegraphics[width=\columnwidth]{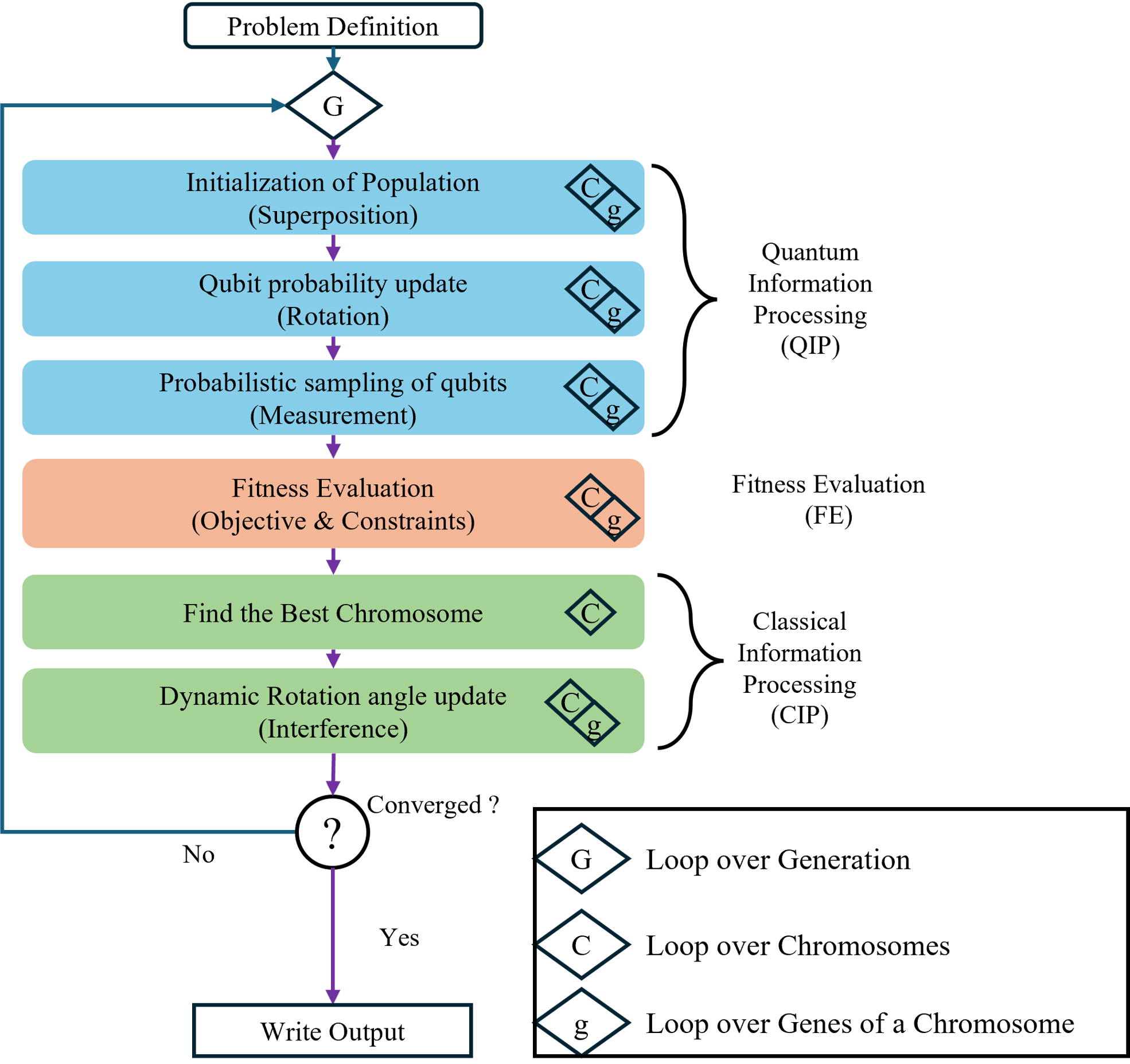}
\caption{QIEO loop. Blue: initialize amplitudes in superposition, rotate each qubit toward the current elite~\eqref{eq:rot}, and measure a classical population. Orange: evaluate. Green: identify the elite chromosome and update the rotation (interference). The loop is a repeated, coordinate-wise pull onto one elite. Diamonds mark chromosome-level (C) versus gene-level (g) operations.}
\label{fig:QIEO_Schematic}
\end{figure}

The QIEO procedure can be summarized as follows (Fig.~\ref{fig:QIEO_Schematic}):

\begin{algorithm}[t]
\caption{Quantum-Inspired Evolutionary Optimization}
\label{alg:qieo}
\begin{algorithmic}[1]
\Require Population size \(N_{\mathrm{pop}}\), maximum generations \(G_{\max}\), rotation angle magnitude \(\left|\Delta\theta\right|\), and domain bounds \(\Omega\)
\State Initialize all qubits using \(\alpha_{j,i}=\beta_{j,i}=1/\sqrt{2}\)
\State Observe the initial quantum-inspired population to create \(\mathcal{P}^{(0)}\)
\State Decode and evaluate \(\mathcal{P}^{(0)}\)
\State Store the best observed chromosome as \(\mathbf{b}_{\mathrm{best}}^{(0)}\)
\For{\(g=1,\ldots,G_{\max}\)}
    \State Determine the sign of \(\Delta\theta\) for every qubit using the elite chromosome
    \State Update all qubit amplitudes using Eq.~\eqref{eq:rot}
    \State Observe the updated quantum-inspired population
    \State Decode the observed chromosomes into design vectors
    \State Evaluate the objective function for all candidates
    \State Update \(\mathbf{b}_{\mathrm{best}}^{(g)}\) if an improved solution is found
    \If{the stopping criterion is satisfied}
        \State \textbf{break}
    \EndIf
\EndFor
\Ensure Best observed solution \(\mathbf{b}_{\mathrm{best}}\)
\end{algorithmic}
\end{algorithm}

The present study utilizes BQP's BQPhy optimizer that is provided as a python library and can be obtained from BQP's business team (please reach out to bizdev@bqpsim.com) under their ongoing super-user program.\footnote{The above discussion pertains to fixed $\Delta \theta$ QIEO variant that utilizes binary encoding. Discussions related to the other variants, viz. adaptive $\Delta \theta$ and real encoded QIEO are not provided as they constitute BQP's Intellectual Property.}

\subsection{Genetic Algorithm}
\label{sec:ga}

A Genetic Algorithm (GA) is a population-based, stochastic optimization method inspired by evolutionary processes. Unlike gradient-based optimizers, GAs do not require derivative information and can be applied to nonlinear, discontinuous, multimodal, and black-box objective functions. A GA evolves a population of candidate solutions by iteratively applying selection, crossover, mutation, and elitism, thereby balancing the exploitation of high-quality candidates with exploration of previously unvisited regions of the search space.

Let the optimization problem be expressed as
\begin{equation}
    \min_{\mathbf{x} \in \Omega} f(\mathbf{x}),
    \label{eq:ga_optimization_problem}
\end{equation}
where \(f(\mathbf{x})\) is the objective function and \(\Omega \subseteq \mathbb{R}^{D}\) is a bounded \(D\)-dimensional search domain. For a maximization-oriented implementation, the objective may equivalently be represented through a fitness function \(F(\mathbf{x})\), defined such that higher fitness corresponds to better candidate solutions. In this work, the GA baseline is implemented using the PyGAD framework and is configured to match the QIEO experiments as closely as permitted by the two algorithmic representations.

Two candidate encodings are considered: a real-valued encoding and a binary encoding. In the real-coded GA, an individual directly represents the optimization variables:
\begin{equation}
    \mathbf{x}
    =
    \left[
        x_1, x_2, \ldots, x_D
    \right]^{\mathsf{T}},
    \qquad
    x_j \in [l_j,u_j],
    \label{eq:real_ga_encoding}
\end{equation}
where \(l_j\) and \(u_j\) denote the lower and upper bounds of the \(j\)-th decision variable, respectively. Therefore, the genes are the physical coordinates themselves, and every individual is constrained to remain within the prescribed box domain.

For the binary-coded GA, each decision variable is represented by a fixed-length binary string of \(16\) bits. Consequently, a \(D\)-dimensional design vector is encoded using
\begin{equation}
    N_{\mathrm{genes}}
    =
    16D
    \label{eq:binary_ga_gene_count}
\end{equation}
binary genes. Let \(\mathbf{b}_j \in \{0,1\}^{16}\) denote the bit string associated with the \(j\)-th variable. Its corresponding unsigned integer representation is
\begin{equation}
    q_j
    =
    \sum_{r=0}^{15}
    b_{j,r}2^r,
    \qquad
    q_j \in \left[0,2^{16}-1\right].
    \label{eq:binary_integer_mapping}
\end{equation}
The binary representation is decoded to the physical coordinate using the same fixed-point mapping employed in the binary QIEO formulation:
\begin{equation}
    x_j
    =
    l_j
    +
    \frac{q_j}{2^{16}-1}
    \left(u_j-l_j\right).
    \label{eq:binary_ga_decoding}
\end{equation}
This common decoding rule ensures that the binary GA and binary QIEO operate over the same discretized domain and have the same per-variable representational resolution.

At each generation, parents are selected using tournament selection with tournament size
\begin{equation}
    K = 3.
\end{equation}
For each parent-selection event, \(K\) individuals are sampled from the population and the individual with the best fitness is selected as a parent. Tournament selection creates selection pressure toward high-fitness candidates while retaining stochasticity, which helps preserve population diversity and reduces the probability of immediate convergence to a suboptimal region.

Selected parent pairs produce offspring through single-point crossover. For a chromosome of length \(L\), a crossover point \(c\) is selected such that
\begin{equation}
    1 \leq c < L.
\end{equation}
Given parent chromosomes \(\mathbf{g}^{(1)}\) and \(\mathbf{g}^{(2)}\), the offspring are formed by exchanging the chromosome segments following the selected crossover point:
\begin{equation}
    \mathbf{o}^{(1)}
    =
    \left[
        g_1^{(1)},\ldots,g_c^{(1)},
        g_{c+1}^{(2)},\ldots,g_L^{(2)}
    \right],
\end{equation}
\begin{equation}
    \mathbf{o}^{(2)}
    =
    \left[
        g_1^{(2)},\ldots,g_c^{(2)},
        g_{c+1}^{(1)},\ldots,g_L^{(1)}
    \right].
\end{equation}

After crossover, random mutation is applied to \(15\%\) of the genes. For the binary representation, mutation changes the selected bit values according to
\begin{equation}
    b_{j,r}
    \leftarrow
    1-b_{j,r}.
    \label{eq:binary_mutation}
\end{equation}
For the real-coded representation, mutation perturbs the selected coordinate genes while enforcing the associated bound constraints:
\begin{equation}
    x_j \in [l_j,u_j].
\end{equation}
The relatively high mutation fraction promotes exploration and reduces the likelihood of population collapse around a locally optimal configuration.

\begin{figure}[!t]
\centering
\includegraphics[width=\columnwidth]{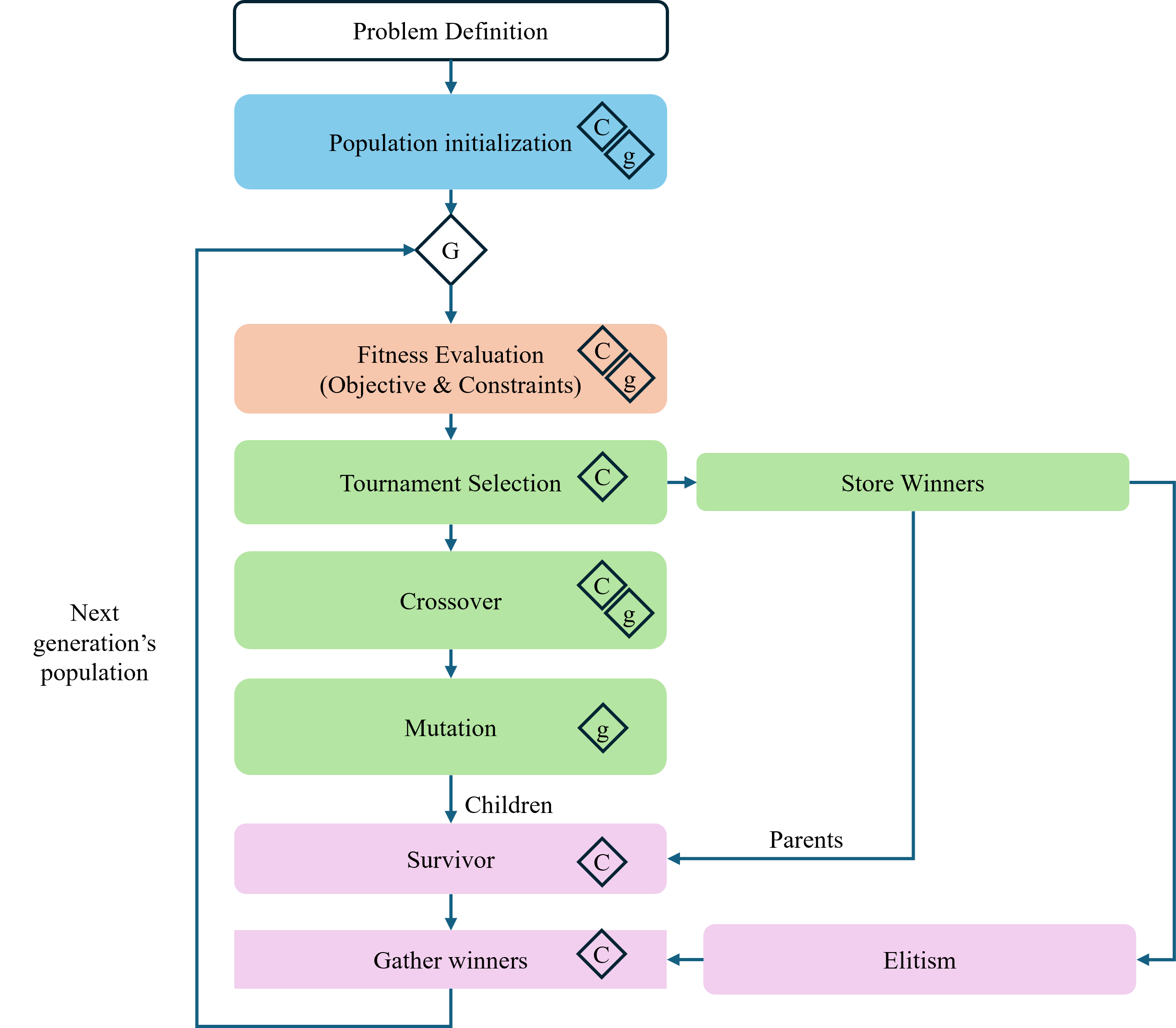}
\caption{GA loop used as the matched-budget baseline. Blue: initialize the population. Orange: evaluate. Green: tournament selection (\(K=3\)), single-point crossover, and gene-wise mutation (\(15\%\)). Pink: survivors, with the two elites copied unchanged (``Store Winners''). Diamonds mark chromosome-level (C) versus gene-level (g) operations.}
\label{fig:GA_Schematic}
\end{figure}

Elitism is used to preserve the two best individuals from each generation without modification. The two highest-fitness individuals are copied directly into the next generation:
\begin{equation}
    N_{\mathrm{elite}} = 2.
\end{equation}
This mechanism guarantees that the best solution identified so far is not lost due to crossover or mutation. Fig.~\ref{fig:GA_Schematic} is the corresponding loop.

\begin{table}[t]
    \centering
    \caption{Genetic Algorithm configuration used as the optimization baseline.}
    \label{tab:ga_configuration}
    \begin{tabular}{ll}
        \toprule
        \textbf{Parameter} & \textbf{Configuration} \\
        \midrule
        Implementation & PyGAD \\
        Population size & \(1000\) \\
        Maximum generations & Same cap as QIEO \\
        Real-coded representation & \(D\) bounded coordinate genes \\
        Binary-coded representation & \(16D\) binary genes \\
        Binary precision & \(16\) bits per decision variable \\
        Selection method & Tournament selection \\
        Tournament size & \(K=3\) \\
        Crossover operator & Single-point crossover \\
        Mutation & Random mutation of \(15\%\) of genes \\
        Elitism & Best two individuals retained unchanged \\
        Early-stopping criterion & No best-fitness improvement for \(40\) generations \\
        \bottomrule
    \end{tabular}
\end{table}

\subsection{Covariance matrix adaptation evolution strategy}

The Covariance Matrix Adaptation Evolution Strategy (CMA-ES) is a stochastic, derivative-free optimization method for continuous decision variables. It is particularly suitable for nonlinear, nonconvex, ill-conditioned, and black-box optimization problems in which analytical gradients are unavailable, inaccurate, noisy, or prohibitively expensive to compute. CMA-ES belongs to the class of evolution strategies and models the search process through an adaptive multivariate Gaussian distribution.

Let the objective be
\begin{equation}
    \min_{\mathbf{x} \in \mathbb{R}^{n}} f(\mathbf{x}),
\end{equation}
where \(f:\mathbb{R}^{n}\rightarrow\mathbb{R}\) is a black-box objective function and \(n\) is the dimension of the design space. At generation \(g\), CMA-ES characterizes its search distribution using a mean vector \(\mathbf{m}^{(g)}\), a global step size \(\sigma^{(g)} > 0\), and a covariance matrix \(\mathbf{C}^{(g)} \in \mathbb{R}^{n \times n}\). Candidate solutions are sampled according to
\begin{equation}
    \mathbf{x}_{k}^{(g)}
    =
    \mathbf{m}^{(g)}
    +
    \sigma^{(g)} \mathbf{y}_{k}^{(g)},
    \qquad
    \mathbf{y}_{k}^{(g)}
    \sim
    \mathcal{N}\left(\mathbf{0},\mathbf{C}^{(g)}\right),
    \qquad
    k=1,\ldots,\lambda,
    \label{eq:cmaes_sampling}
\end{equation}
where \(\lambda\) denotes the offspring population size.

The sampled offspring are evaluated and ranked according to their objective values:

\begin{equation}
    f\left(\mathbf{x}_{1:\lambda}^{(g)}\right)
    \leq
    f\left(\mathbf{x}_{2:\lambda}^{(g)}\right)
    \leq
    \cdots
    \leq
    f\left(\mathbf{x}_{\lambda:\lambda}^{(g)}\right),
\end{equation}
where \(\mathbf{x}_{i:\lambda}^{(g)}\) denotes the \(i\)-th best candidate among the \(\lambda\) offspring. CMA-ES is rank-based: the selection mechanism depends on the ordering of objective values rather than their absolute magnitudes.

The distribution mean is updated by weighted recombination of the best \(\mu\) offspring:
\begin{equation}
    \mathbf{m}^{(g+1)}
    =
    \sum_{i=1}^{\mu}
    w_i \mathbf{x}_{i:\lambda}^{(g)},
    \qquad
    \sum_{i=1}^{\mu} w_i = 1,
    \qquad
    w_i > 0,
    \label{eq:cmaes_mean_update}
\end{equation}
where the recombination weights \(w_i\) are generally chosen such that higher-ranked offspring receive greater weight. The effective selection mass is defined as
\begin{equation}
    \mu_{\mathrm{eff}}
    =
    \left(
        \sum_{i=1}^{\mu} w_i^2
    \right)^{-1}.
\end{equation}

The main distinguishing feature of CMA-ES is the adaptation of the covariance matrix. Rather than sampling isotropically in all coordinate directions, CMA-ES learns the scale and correlation structure of successful search steps. The resulting search distribution can become elongated and rotated, thereby aligning with valleys, ridges, and coupled parameter directions in the objective landscape.

Two evolution paths are used in the standard algorithm. The step-size evolution path is given by
\begin{equation}
    \mathbf{p}_{\sigma}^{(g+1)}
    =
    (1-c_{\sigma})\mathbf{p}_{\sigma}^{(g)}
    +
    \sqrt{c_{\sigma}(2-c_{\sigma})\mu_{\mathrm{eff}}}
    \left(\mathbf{C}^{(g)}\right)^{-1/2}
    \frac{
        \mathbf{m}^{(g+1)}-\mathbf{m}^{(g)}
    }{
        \sigma^{(g)}
    },
    \label{eq:step_size_path}
\end{equation}
where \(c_{\sigma}\) is the step-size path learning rate. This path is used in cumulative step-size adaptation (CSA):
\begin{equation}
    \sigma^{(g+1)}
    =
    \sigma^{(g)}
    \exp
    \left[
        \frac{c_{\sigma}}{d_{\sigma}}
        \left(
            \frac{
                \left\|
                \mathbf{p}_{\sigma}^{(g+1)}
                \right\|
            }{
                \mathbb{E}\left[
                    \left\|
                    \mathcal{N}(\mathbf{0},\mathbf{I})
                    \right\|
                \right]
            }
            -1
        \right)
    \right],
    \label{eq:step_size_update}
\end{equation}
where \(d_{\sigma}\) is a damping parameter. Persistent movement in similar directions produces an evolution path longer than expected under random sampling, leading to an increase in the global step size. Conversely, uncorrelated or oscillatory progress tends to reduce the step size.

The covariance evolution path is updated as
\begin{equation}
    \mathbf{p}_{c}^{(g+1)}
    =
    (1-c_c)\mathbf{p}_{c}^{(g)}
    +
    h_{\sigma}
    \sqrt{c_c(2-c_c)\mu_{\mathrm{eff}}}
    \frac{
        \mathbf{m}^{(g+1)}-\mathbf{m}^{(g)}
    }{
        \sigma^{(g)}
    },
    \label{eq:covariance_path}
\end{equation}
where \(c_c\) is the covariance path learning rate and \(h_{\sigma}\) is an indicator controlling the contribution of the path update. The covariance matrix is then updated using rank-one and rank-\(\mu\) contributions:
\begin{equation}
\begin{split}
    \mathbf{C}^{(g+1)}
    ={}&
    \left(1-c_1-c_{\mu}\right)
    \mathbf{C}^{(g)}
    +
    c_1
    \mathbf{p}_{c}^{(g+1)}
    \left(\mathbf{p}_{c}^{(g+1)}\right)^{\mathsf{T}}
    \\
    &+
    c_{\mu}
    \sum_{i=1}^{\mu}
    w_i
    \mathbf{y}_{i:\lambda}^{(g)}
    \left(\mathbf{y}_{i:\lambda}^{(g)}\right)^{\mathsf{T}},
    \label{eq:covariance_update}
\end{split}
\end{equation}
where \(c_1\) and \(c_{\mu}\) are the learning rates for the rank-one and rank-\(\mu\) updates, respectively. The rank-one term captures persistent search directions over multiple generations, whereas the rank-\(\mu\) term incorporates the distribution of successful offspring within the current generation.

Initially, the covariance matrix is commonly initialized as an identity matrix,
\begin{equation}
    \mathbf{C}^{(0)} = \mathbf{I},
\end{equation}
which corresponds to isotropic sampling:
\begin{equation}
    \mathbf{x}
    \sim
    \mathcal{N}
    \left(
        \mathbf{m},
        \sigma^2 \mathbf{I}
    \right).
\end{equation}
As optimization proceeds, covariance adaptation produces a general anisotropic Gaussian distribution,
\begin{equation}
    \mathbf{x}
    \sim
    \mathcal{N}
    \left(
        \mathbf{m},
        \sigma^2 \mathbf{C}
    \right).
\end{equation}

\begin{figure}[!t]
\centering
\includegraphics[width=\columnwidth]{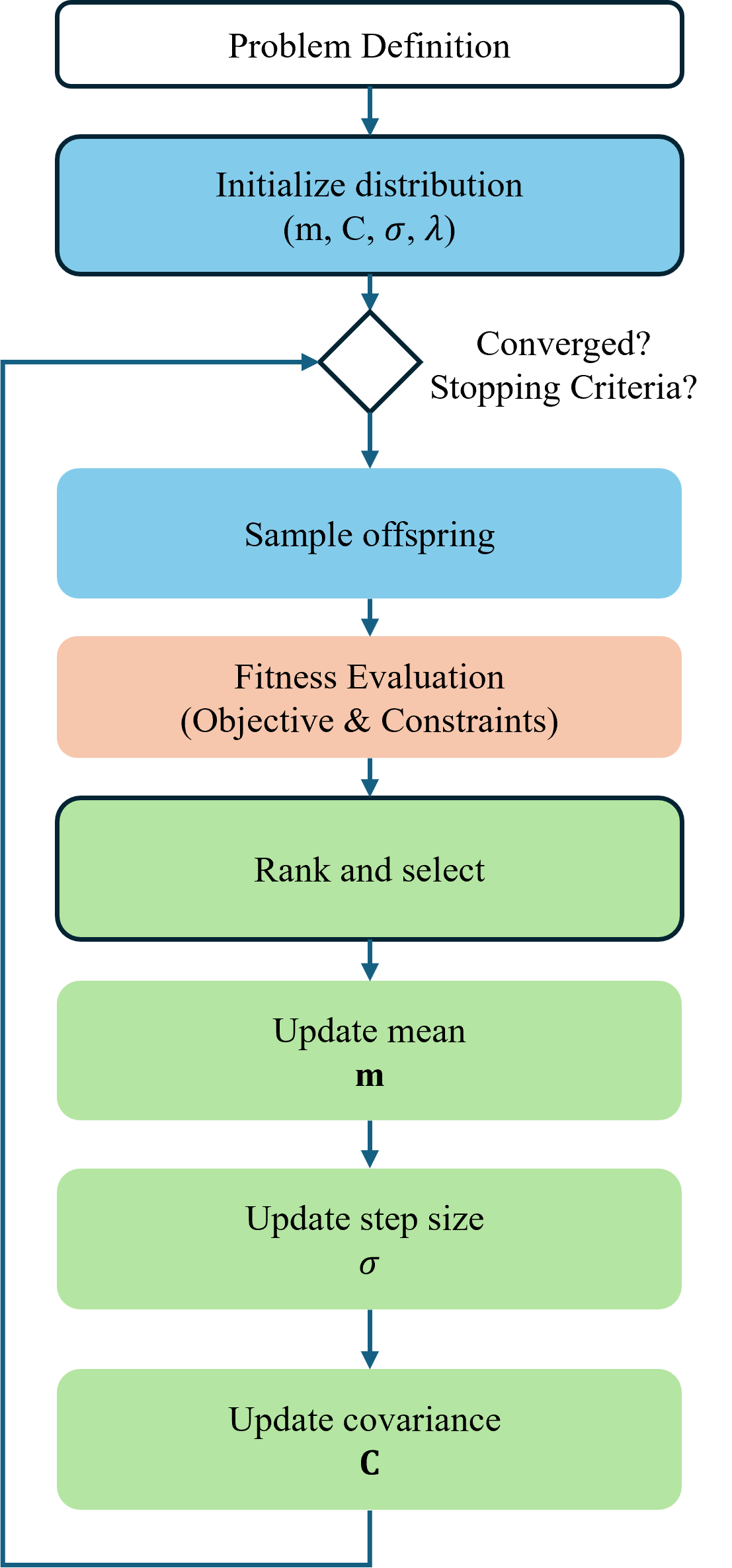}
\caption{CMA-ES loop (Hansen default, no restarts). Blue: initialize \((\mathbf{m},\mathbf{C},\sigma,\lambda)\) and sample \(\lambda\) offspring from \(\mathcal{N}(\mathbf{m},\sigma^{2}\mathbf{C})\). Orange: evaluate. Green: rank, then update the mean, the global step size \(\sigma\) (CSA), and the covariance \(\mathbf{C}\) (rank-one and rank-\(\mu\)). The object that evolves is the search distribution, not a set of chromosomes.}
\label{fig:CMA_Schematic}
\end{figure}

The eigenvectors of \(\mathbf{C}\) identify learned search directions, whereas its eigenvalues determine the exploration scale along those directions. This mechanism enables CMA-ES to adapt to rotated and ill-conditioned landscapes, including problems with strongly correlated decision variables. Such behavior is valuable in simulation-driven optimization, where physically meaningful parameters often exhibit substantial coupling. The learned \(C\) is what lets the method follow a narrow, rotated, ill-conditioned valley, and it is precisely the object that Eq.~\eqref{eq:rot} cannot express. Fig.~\ref{fig:CMA_Schematic} is the loop: no chromosome population is retained, only the Gaussian \((\mathbf{m},\sigma,\mathbf{C})\).

This research uses Hansen's \texttt{pycma} CMA-ES implementation~\cite{hansen2016tutorial}. We use the textbook defaults throughout: population \(\lambda=4+\lfloor 3\ln D\rfloor\), initial mean at the box centre, \(\sigma_0=0.3\) times the median side length, bound handling by the solver, and restarts disabled. Note how small that population is next to QIEO's---six candidates per generation at \(D=2\) against one thousand---which is the single most important asymmetry to keep in mind when reading the results. CMA-ES also stops on its own internal tolerances on step size, objective spread, and stagnation rather than on the shared cap.

Although CMA-ES can efficiently adapt its local search distribution, a single run may converge prematurely to a local optimum on a multimodal objective landscape. Restart strategies address this limitation by reinitializing CMA-ES after stagnation, convergence to an undesired basin, covariance degeneracy, or exhaustion of a prescribed evaluation budget.

We deliberately use the textbook default rather than IPOP or BIPOP, because the comparison of interest is against the solver a practitioner actually reaches for first. Section~\ref{sec:disc} explains why this choice matters for interpreting our cost figures.

\section{Experimental Setup}
\subsection{Objectives and their labels}

Let \(\mathcal{F}\) be the 256 continuous functions implemented in this research\footnote{the code for these function can be found in \texttt{index.py} along with the box bounds and dimension \(D\) which are recorded in \texttt{configs.py} within the attached replication package}. Every \(f\in\mathcal{F}\) carries the eleven labels of Table~\ref{tab:traits}, taken from Jamil and Yang~\cite{jamil2013survey}, from Gavana's~\cite{gavana2013} class attributes, and where those disagree or are silent, from inspection of the coded formula. Labels were assigned before the campaigns were analysed and never from optimizer traces, so they cannot have been fitted to the outcomes.

\begin{table}[!t]
\caption{Landscape labels used as predictors of strict success.}
\label{tab:traits}
\centering
\begin{tabular}{ll}
\toprule
Label & Levels used in the analysis \\
\midrule
Continuity & Continuous, Discontinuous \\
Differentiability & Differentiable, Non-differentiable \\
Separability & Separable, Partially separable, Non-separable \\
Scalability & Scalable, Non-scalable \\
Modality & Unimodal, Multimodal \\
Convexity & Convex, Non-convex \\
Conditioning & Well-conditioned, Ill-conditioned \\
Symmetry & Symmetric, Asymmetric \\
Maximum dimensionality & Native 2-D, 3--5-D, \(\geq\)6-D, Scalable \(D\) \\
\(f^\star(D)\) & Yes / No (analytic minimum depends on \(D\)) \\
Variable coupling & None, Adjacent pairs, Fixed-\(D\), All-to-all \\
\bottomrule
\end{tabular}
\end{table}

\begin{enumerate}
\item \textbf{Continuity:} A landscape is continuous when arbitrarily small perturbations in the decision vector produce arbitrarily small changes in the objective value. Discontinuities create jumps, steps, or isolated changes in objective value, weakening the reliability of local interpolation and gradient-based search.

\item \textbf{Differentiability:} Differentiability denotes the existence of well-defined derivatives, and hence local directional information, over the relevant search domain. Non-differentiable points, such as kinks and cusps, obstruct conventional gradient or Hessian approximations and favor derivative-free search strategies.

\item \textbf{Separability:} A function is separable when it can be decomposed into independent single-variable contributions, allowing each coordinate to be optimized independently. Non-separability requires coordinated changes across variables and is consequently more challenging for coordinate-wise or independently sampled optimizers.

\item \textbf{Scalability:} Scalability describes whether a benchmark can be extended to higher dimensionality while retaining its defining structural characteristics. It tests whether an optimizer maintains efficiency and solution quality as the search-space dimension and computational burden increase.

\item \textbf{Modality:} Modality is the number and distribution of local extrema within the search landscape. Multimodal landscapes challenge an optimizer to maintain sufficient exploration to avoid premature convergence to suboptimal local minima.

\item \textbf{Convexity:} A convex minimization landscape satisfies the property that every local minimum is globally optimal. Non-convexity introduces local minima, ridges, saddles, and disconnected basins of attraction, making global rather than merely local search necessary.

\item \textbf{Conditioning:} Conditioning measures the disparity in objective sensitivity along different directions, often associated locally with the ratio of the largest to smallest curvature. Ill-conditioned landscapes contain narrow valleys or strongly unequal scales, making progress slow or unstable unless step sizes and search directions adapt appropriately.

\item \textbf{Symmetry:} Symmetry refers to invariance of the objective under one or more transformations of the decision vector, such as sign reversal, permutation, translation, or rotation. Symmetry may create multiple equivalent optima and redundant search regions, thereby complicating assessment of convergence location while not necessarily altering the optimal objective value.

\item \textbf{Maximum dimensionality:} Maximum dimensionality. Maximum dimensionality denotes the largest number of decision variables $D$ for which a benchmark function is defined or can be meaningfully instantiated. Functions with an unbounded or arbitrarily scalable dimension are suitable for evaluating high-dimensional optimization performance, whereas fixed-dimensional functions primarily assess behavior on problems of limited size

\item \textbf{$f^\star(D)$ dependency on dimension:} This property specifies whether the global optimal objective value $f^\star(D)$ varies with the problem dimension $D$. A dimension-dependent optimum requires performance measures to account for the changing target value; otherwise, raw objective values may not be comparable across dimensionalities.

\item \textbf{Variable coupling:} Variable coupling is the degree to which the contribution or optimum of one variable depends on the values of other variables. Strong coupling produces curved valleys and correlated search directions, challenging methods that use independent marginal distributions or uncoordinated perturbations; fully non-separable problems exhibit interactions among every pair of decision variables.

\end{enumerate}

Coupling deserves comment because it is easy to mistake for a synonym of non-separability. We distinguish a sliding pairwise residual, in which coordinate \(i\) interacts only with \(i+1\) as in Rosenbrock or Dixon--Price, from an aggregate form in which a single sum or product mixes every coordinate at once, from a low-dimensional polynomial whose cross terms exist only because the function is defined at a fixed \(D\). All three are non-separable, but they present very different geometry to an operator that updates one coordinate at a time, and Section~\ref{sec:couple} shows they behave differently in the data.

\subsection{Shifted landscapes}
\label{sec:shift-def}

Algebraic benchmarks often place \(x^\star\) at a distinguished location: the origin, the box centre, or a bound. A coordinate-wise sampler that observes each axis independently can meet those points by accident, which would inflate success for QIEO and a GA without reflecting any ability to follow the landscape. The shifted campaign is a control for that artefact. It is the same coded formula, the same box, and the same published \(f^\star\); only the location of the minimizer inside the box is changed.

Let \(f\) be defined on the original box \([\ell,u]\) with a tabulated interior minimizer \(x^\star\). A relocation \(x_{\mathrm{new}}\) is drawn once per function from the uniform distribution on \([\ell,u]\), with the draw seeded by a hash of the function's name so that every encoding and every solver sees the same shift. The optimizer still proposes points \(x\in[\ell,u]\). The evaluated shift is expressed as,
\begin{equation}
\tilde f(x)
=
f\!\bigl(x-x_{\mathrm{new}}+x^\star\bigr)
+
P(x),
\label{eq:shift}
\end{equation}
where \(z=x-x_{\mathrm{new}}+x^\star\) is the preimage in the original coordinates and \(P(x)\) is a quadratic penalty on any coordinate of \(z\) that leaves \([\ell,u]\), plus a large barrier if that occurs. Consequently \(\tilde f(x_{\mathrm{new}})=f^\star\) and \(P(x_{\mathrm{new}})=0\), while a search that collapses to the origin, the centre, or a bound is no longer rewarded unless that point happens to be \(x_{\mathrm{new}}\).

Equation~\eqref{eq:shift} can be formed only when a single \(x^\star\) is known and already lies in \([\ell,u]\). Five names fail that test and are omitted from the shifted comparison, leaving \(n=251\). Lennard-Jones and ZeroSum have no unique tabulated coordinate vector (a cluster geometry and the hyperplane \(\sum_i x_i=0\), respectively). Meyer, Meyer--Roth, and Osborne have published parameter vectors that fall outside the coded box, so a translation that preserves \(f^\star\) at an interior \(x_{\mathrm{new}}\) is not well-defined on those bounds. Shifted results are therefore a paired robustness check on 251 landscapes, not a second independent suite of 256 functions.

\subsection{Termination conditions}

For QIEO and GA, optimization terminates when the maximum generation cap is reached or when the best fitness exhibits no improvement for a prescribed number of consecutive generations. Let \(f_{\mathrm{best}}^{(g)}\) denote the best objective value observed up to generation \(g\). For a stagnation window of \(N_{\mathrm{stall} = 40}\) generations, early termination is applied when
\begin{equation}
    f_{\mathrm{best}}^{(g)}
    =
    f_{\mathrm{best}}^{(g-1)}
    =
    \cdots
    =
    f_{\mathrm{best}}^{(g-N_{\mathrm{stall}}+1)}.
    \label{eq:qieo_early_stopping}
\end{equation}

Using the same population size, domain bounds, generation cap, fixed-point decoding rule, and early-stopping criterion as the corresponding GA baseline permits a controlled comparison of the two search mechanisms. In this setting, the key algorithmic distinction is that the GA alters explicit classical chromosomes through selection, crossover, and mutation, whereas QIEO updates a factored probability representation by repeated rotation toward a single elite solution.
We run three variants. 
\begin{enumerate}
\item \emph{Fixed \(\theta\)} uses a constant step \(\Delta\theta=0.05\) with the binary encoding. 

\item \emph{Adaptive \(\theta\)} uses the same encoding with a magnitude scheduled internally by the implementation from the generation index and the elite--individual disagreement.

\item \emph{Real-coded} replaces the bit decoding with a direct amplitude-to-coordinate map governed by a balance parameter \(\lambda=0.5\) that trades exploration against exploitation.
\end{enumerate}

All three use \(N_p=1000\) individuals for at most \(G_{\max}=5000\) generations and all are taken from the BQPhy Optimiser implementation described in~\cite{eswara2024qieo}. 

The production campaign carries an early-stopping rule that activates if the best individual's fitness remains unchanged (with specified tolerance of $1e-6$) for 40 consecutive generations. 

The study repeated all three variants over all \(256\) and Section~\ref{sec:cost} measures the required generations as QIEO cost.

Although both algorithms are assigned the same nominal generation cap, their actual number of executed generations and objective-function evaluations can be substantially smaller. This explains why the measured computational expenditure reported in Section~\ref{sec:cost} is below the maximum budget implied by the nominal generation limit.

\subsection{Implications of the imposed budget cap}

It is essential to be clear about what a shared \emph{cap} does and does not equalize. QIEO and the GA are population methods configured to that population and generation count, but neither actually consumes what it is permitted: the GA halts on its own stagnation rule, and QIEO in Section~\ref{sec:cost} settles at a median of \(\qieoMedEv\) evaluations, some \(\qieoCapFrac\%\) of the cap. 

CMA-ES with default \(\lambda\) and no restarts stops on its internal tolerances long before the cap, typically after a few hundred evaluations. 

A shared cap therefore equalizes permissible limit, not expenditure, and Section~\ref{sec:cost} treats the resulting asymmetry as a finding rather than a nuisance.

\subsection{Scoring}
\label{sec:score}
Let \(\hat f\) be the best coded value a solver returns and \(f^\star\) the published reference. The scaled residual is
\[
e=\lvert\hat f-f^\star\rvert
\quad\text{if}\quad
\lvert f^\star\rvert<10^{-2},
\qquad
e=\lvert\hat f-f^\star\rvert\big/\lvert f^\star\rvert
\quad\text{otherwise}.
\]
Outcomes are then labelled
\begin{itemize}
\item \textbf{Excellent} if \(e<10^{-6}\);
\item \textbf{Good} if \(10^{-6}\le e<10^{-3}\);
\item \textbf{Fair} if \(10^{-3}\le e<5\times10^{-2}\) (absolute) or \(10^{-3}\le e<10^{-1}\) (relative);
\item \textbf{Poor} if \(e\) meets or exceeds that Fair ceiling;
\item \textbf{Bad} if \(\hat f\) undercuts \(f^\star\) by more than the Good slack (\(10^{-3}\) absolute, \(10^{-3}\lvert f^\star\rvert\) relative), which signals an implementation discrepancy or an incorrect published reference.
\end{itemize}

Another set of qualifiers that need to be defined are, 
\begin{itemize}
\item \emph{Strict success} means Excellent or Good, which corresponds to a scaled residual below roughly \(10^{-3}\). The \(10^{-3}\) cutoff that defines strict success is a choice, not a property of the solvers. Section~\ref{sec:precision} therefore reports the distribution of scaled residuals, not only the pass rate, and repeats the ranking at several tolerances.

\item ``QIEO best-of-3'' and ``GA best-of-2'' denote the better outcome across encodings of that family; they are \emph{oracle} upper bounds on what a user could obtain by trying every encoding, not single runs.
\end{itemize}

\subsection{Statistical Analysis}

For a subset of \(n\) functions of which \(k\) are strict successes, we report
\[
\hat{p}=k/n
\]
as a percentage \(100\hat{p}\), together with a two-sided \(95\%\) Wilson score interval~\cite{wilson1927,brown2001}. Wilson intervals are used instead of the normal (Wald) approximation because they have more reliable coverage for the small and unbalanced strata produced by labelling, especially when \(\hat{p}\) is near \(0\) or \(1\). These intervals appear on the aggregate rates in Table~\ref{tab:overall} and Fig.~\ref{fig:overall}; the same \(\hat{p}=k/n\) is the quantity displayed by characteristic in Fig.~\ref{fig:radar} and against dimension in Fig.~\ref{fig:dim}.

For each of the nine binary labels in Table~\ref{tab:traits}, association with real-coded QIEO strict success is assessed from the \(2\times 2\) table of (success, failure) against (level A, level B). Independence is tested by Pearson's \(\chi^2\) with Yates' continuity correction~\cite{yates1934}. We report the \(p\)-value together with an odds ratio and its interval, so that the size and direction of the association are visible separately from significance. The nine contrasts, odds ratios, and \(p\)-values are collected in Table~\ref{tab:traits2} and discussed in Section~\ref{sec:props}.

To keep the odds ratio defined on sparse tables, including those with a zero cell, we apply the Haldane--Anscombe correction, adding \(0.5\) to every cell~\cite{haldane1956,anscombe1956}. For
\[
\begin{array}{c|cc}
 & \text{Strict success} & \text{Failure} \\
\hline
\text{Level A} & a & b \\
\text{Level B} & c & d
\end{array},
\]
the corrected odds ratio is
\[
\widehat{\mathrm{OR}}
=
\frac{(a+0.5)(d+0.5)}{(b+0.5)(c+0.5)}.
\]
Its interval is the Woolf interval on the log scale~\cite{woolf1955}, using the same adjusted counts. \(\widehat{\mathrm{OR}}>1\) means higher estimated odds of strict success at Level A than at Level B.

Because the nine tests share one outcome, Table~\ref{tab:traits2} reports both raw and Holm-adjusted \(p\)-values~\cite{holm1979}. Holm's sequentially rejective procedure controls the family-wise error rate and is uniformly more powerful than a single-step Bonferroni correction. Unless otherwise stated, claims that survive multiple comparison use the Holm-adjusted values. Section~\ref{sec:props} shows that this changes which labels are called significant.

The labels in Table~\ref{tab:traits} are strongly correlated, so a univariate contrast partly measures its neighbours. Section~\ref{sec:multi} therefore fits a logistic model~\cite{cox1958} of strict success on the five binary predictors retained in Table~\ref{tab:logit}---multimodality, non-separability, ill-conditioning, asymmetry, and non-differentiability---together with \(\log_2 D\):
\[
\operatorname{logit}\!\bigl[\Pr(Y_i=1)\bigr]
=
\beta_0
+\sum_{j=1}^{5}\beta_j x_{ij}
+\beta_D\log_2(D_i),
\]
where \(Y_i=1\) denotes strict success on function \(i\), \(x_{ij}\in\{0,1\}\) is the \(j\)-th binary trait, and \(D_i\) is that function's dimension. Then \(\mathrm{e}^{\beta_j}\) is the odds ratio for trait \(j\) given the other predictors, and \(\mathrm{e}^{\beta_D}\) is the odds ratio for a doubling of \(D\). The same specification is fitted separately for real-coded QIEO, CMA-ES, and the GA best-of-2 oracle; the three columns of Table~\ref{tab:logit} are those fits. Fig.~\ref{fig:dim} is the corresponding unadjusted view of the dimension term.

Parameters are estimated by iteratively reweighted least squares~\cite{nelder1972}. A small ridge penalty is added to the coefficients for numerical stability when predictors are correlated or nearly separating~\cite{lecessie1992}. The joint model is an adjusted association; it is not a causal claim.

\section{Results}
\subsection{Comparison of aggregate strict-success rates}
\label{Sec:success_rate}
\begin{figure*}[!t]
\centering
\includegraphics[width=\textwidth]{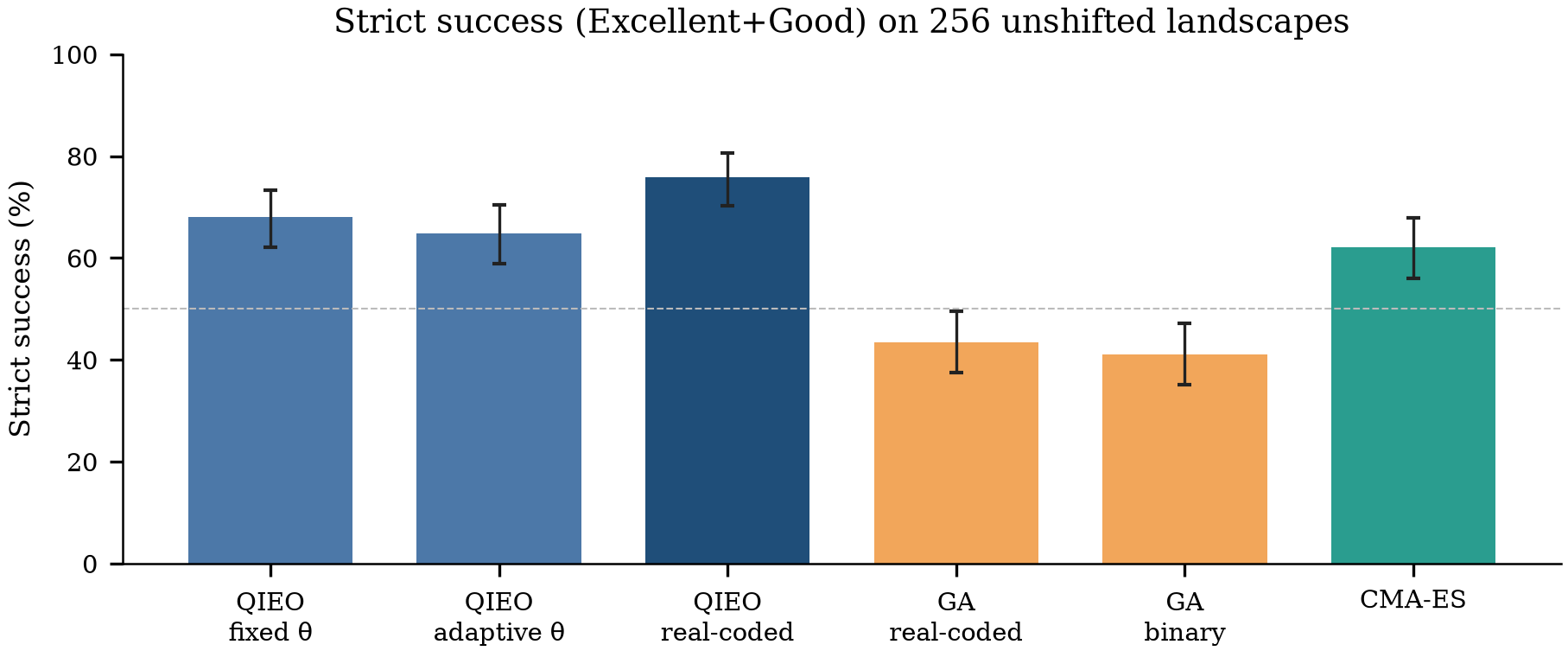}
\caption{Strict success on 256 unshifted names, with Wilson \(95\%\) intervals. Real-coded QIEO leads on raw hit rate; the GA does not approach either QIEO or CMA-ES. Sections~\ref{sec:cost} and~\ref{sec:precision} show why this ordering is not the whole story.}
\label{fig:overall}
\end{figure*}

Table~\ref{tab:overall} and Fig.~\ref{fig:overall} report unshifted strict success. Real-coded QIEO attains \(\QieoRealPct\%\). The binary QIEO encodings trail by eight and eleven points. CMA-ES attains \(\CmaPct\%\)and both GA encodings remain in the low forties. An oracle over the three QIEO encodings raises the rate only to \(\QieoBestPct\%\), so the real-coded variant already accounts for nearly all of the family's successes. The corresponding GA oracle reaches \(\GaBestPct\%\).

\begin{table}[!t]
\caption{Unshifted strict success (Excellent+Good) on \(n=256\) names. Intervals are Wilson \(95\%\). Best-of rows are oracles over encodings, not single runs.}
\label{tab:overall}
\centering
\begin{tabular}{lccc}
\toprule
Solver & \(k/n\) & \% & \(95\%\) CI \\
\midrule
QIEO real-coded & \(\QieoRealK/\QieoRealN\) & \(\QieoRealPct\) & \([\QieoRealLo,\QieoRealHi]\) \\
QIEO best-of-3 (oracle) & \(\QieoBestK/\QieoBestN\) & \(\QieoBestPct\) & \([\QieoBestLo,\QieoBestHi]\) \\
QIEO fixed \(\theta\) & \(\QieoFixK/\QieoFixN\) & \(\QieoFixPct\) & \([\QieoFixLo,\QieoFixHi]\) \\
QIEO adaptive \(\theta\) & \(\QieoAdpK/\QieoAdpN\) & \(\QieoAdpPct\) & \([\QieoAdpLo,\QieoAdpHi]\) \\
CMA-ES (Hansen default) & \(\CmaK/\CmaN\) & \(\CmaPct\) & \([\CmaLo,\CmaHi]\) \\
GA best-of-2 (oracle) & \(\GaBestK/\GaBestN\) & \(\GaBestPct\) & \([\GaBestLo,\GaBestHi]\) \\
GA real-coded & \(\GaRealK/\GaRealN\) & \(\GaRealPct\) & \([\GaRealLo,\GaRealHi]\) \\
GA binary & \(\GaBinK/\GaBinN\) & \(\GaBinPct\) & \([\GaBinLo,\GaBinHi]\) \\
\bottomrule
\end{tabular}
\end{table}

A cautious outlook is advised while considering these results.

This is not evidence that QIEO dominates CMA-ES on BBOB or CEC, where the dimension distribution, the rotations and the accuracy targets are all different. It is evidence that on this 256-function suite, under this budget cap and specific scoring criteria defined in Section \ref{sec:score}, a large-population coordinate-wise rotator locates published minima more often than a default CMA-ES, and that a GA given the identical population and generation budget does not come close. The GA result is the least ambiguous finding in the paper and reveals that the rotation gate is considerably more effective than crossover and mutation with the same sampling allowance.

The rest of this section presents the three caveat that have to be considered while interpreting the hit-rate lead as a ``QIEO wins''.

\subsection{Caveat 1: Evaluation cost}
\label{sec:cost}

\begin{figure}[!t]
\centering
\includegraphics[width=\columnwidth]{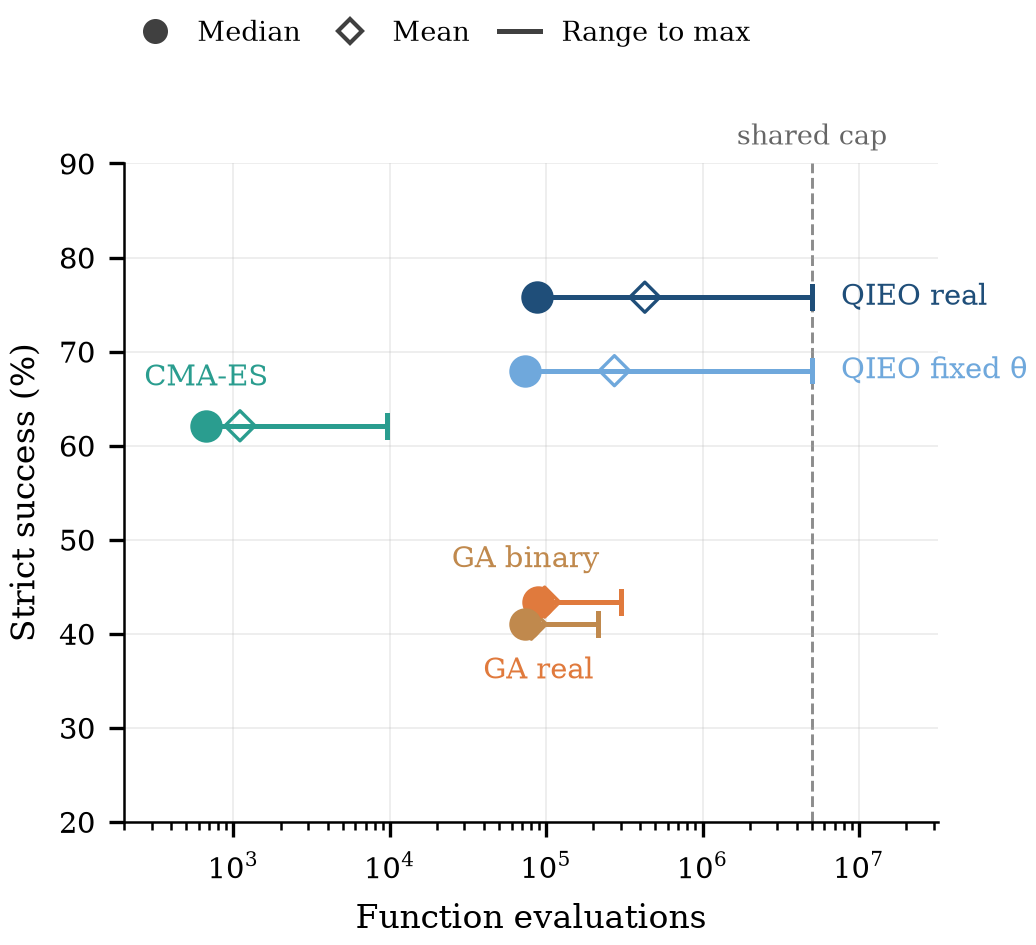}
\caption{Cost against quality. Filled markers are the median evaluations per function; open diamonds are the mean; bars run from the median to the maximum. The dashed line is the shared cap of \(5\times10^{6}\) evaluations. CMA-ES's tail stays far below the cap; QIEO's reaches it. CMA-ES and QIEO occupy the two Pareto-efficient corners; both GA encodings are dominated on cost and quality.}
\label{fig:cost}
\end{figure}

The shared cap is a limit, not expenditure. Fig.~\ref{fig:cost} plots strict success against evaluations actually consumed (median, mean, and range to the maximum). CMA-ES uses a median of \(\cmaMedEv\) evaluations and a maximum of \(\cmaMaxEv\) (\(\cmaCapFrac\%\) of the cap), stopping on its internal tolerances. The GA uses a median of \(\gaMedEv\). QIEO, stopped on the same 40-generation fitness-stagnation rule as the GA, uses a median of \(\qieoMedEv\) evaluations (\(\qieoMedGen\) generations), or \(\qieoCapFrac\%\) of the cap.

The mean for real-coded QIEO is \(\qieoMeanEv\), about five times the median, because \(\qieoAtCap\) of 256 runs reach the cap. Those runs account for most of the campaign total of \(\qieoTotMEv\) million evaluations. Typical cost is tens of thousands of evaluations with only a minority of functions consuming the full budget.

Per million evaluations, CMA-ES solves \(\cmaPerM\) functions and QIEO \(\qieoPerM\): a factor of \(\cmaOverQieo\). Charging QIEO the cap instead of measured cost would still leave a gap of two to three orders of magnitude. The GA's yield of \(\gaPerM\) is inflated by early stagnation on hard functions (total \(\gaTotMEv\) million evaluations versus QIEO's \(\qieoTotMEv\) million). In the median plotted in Fig.~\ref{fig:cost}, the GA costs more than real-coded QIEO and solves fewer functions.

Normalizing turns the ranking inside out. Per million evaluations, CMA-ES solves \(\cmaPerM\) functions and QIEO \(\qieoPerM\), which is a factor of \(\cmaOverQieo\). Charging QIEO the budget cap instead of measured cost would still leave a gap of two to three orders of magnitude. GA's yield of \(\gaPerM\) is inflated by early stagnation on hard functions (total \(\gaTotMEv\) million evaluations versus QIEO's \(\qieoTotMEv\) million). In the median plotted in Fig.~\ref{fig:cost}, the GA costs more than real-coded QIEO and solves fewer functions.

This reframes the aggregate result rather than refuting it, and it does so in a way that is directly useful.If each evaluation is expensive, the hit-rate lead is not worth the extra sampling. If evaluations are cheap and parallel~\cite{eswara2024qieo}, wall-clock time, not evaluation count, is the constraint. CMA-ES and QIEO occupy the two efficient corners of that trade-off; both GA encodings are dominated on cost and on quality.

\subsection{Caveat 2: Degree of accuracy}
\label{sec:precision}

\begin{figure}[!t]
\centering
\includegraphics[width=\columnwidth]{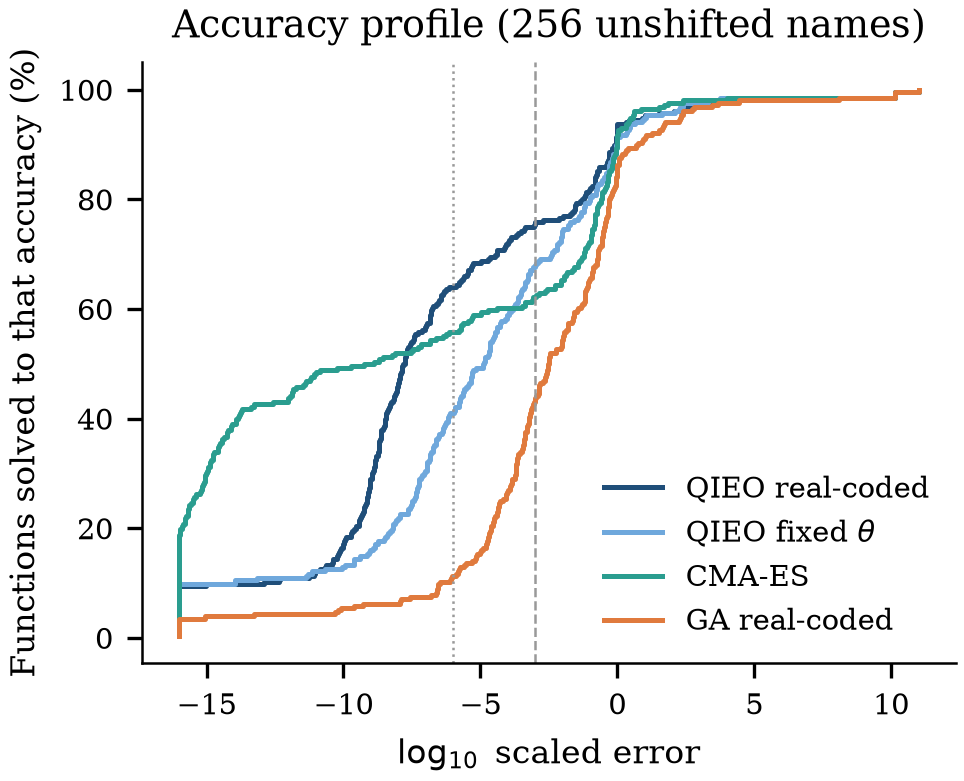}
\caption{Accuracy profiles over 256 unshifted names. The dotted and dashed verticals mark the \(10^{-6}\) and \(10^{-3}\) gates. CMA-ES is nearly bimodal, reaching machine precision or nothing; real-coded QIEO fills the intermediate band that the \(10^{-3}\) gate rewards. The QIEO and CMA-ES curves cross near \(10^{\flipExp}\).}
\label{fig:acc}
\end{figure}

\begin{figure}[!t]
\centering
\includegraphics[width=\columnwidth]{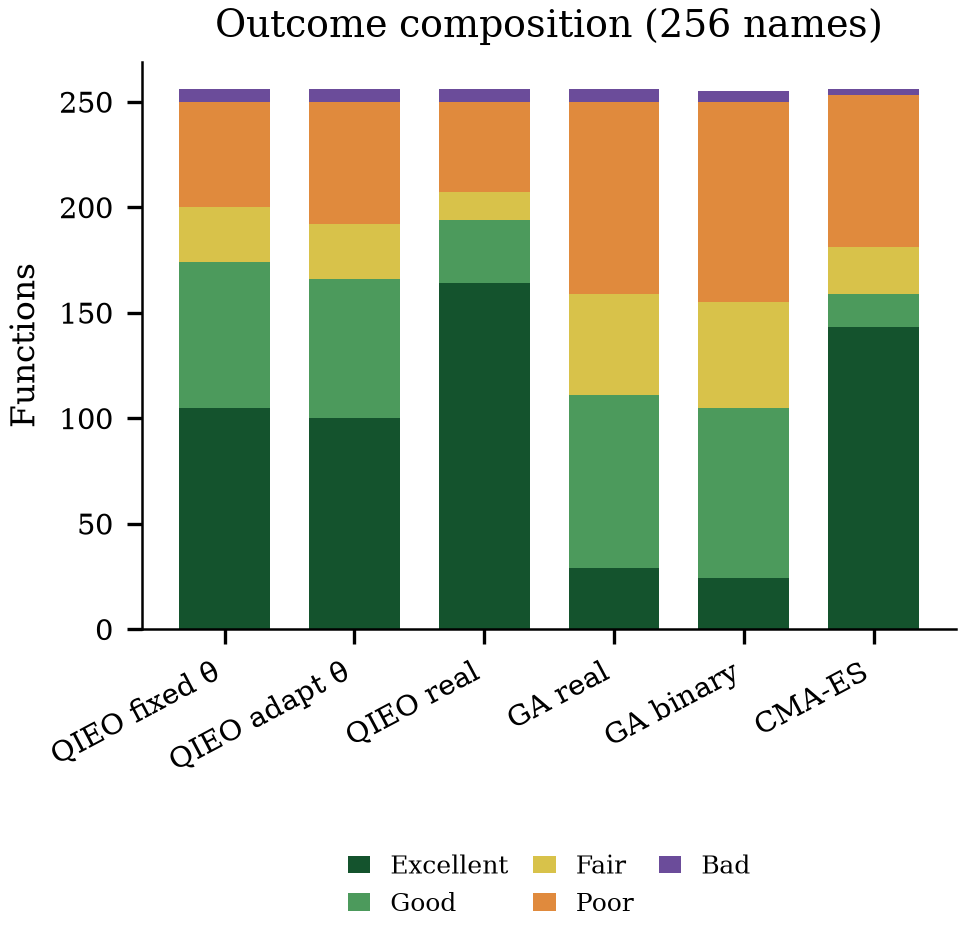}
\caption{Outcome composition for each configuration. CMA-ES converts a smaller number of successes into a higher proportion of Excellent grades; the GA accumulates Good and Fair, rarely converging tightly. The GA is best described as an approximator rather than a solver.}
\label{fig:stack}
\end{figure}

Fig.~\ref{fig:acc} reports the fraction of the suite reaching any given scaled residual for each solver. CMA-ES reaches a residual below \(10^{-12}\) on \(\bandCTwelve\%\) of functions, against \(\bandQTwelve\%\) for real-coded QIEO, and places little mass between \(10^{-8}\) and \(10^{-3}\). Real-coded QIEO does the reverse. Of CMA-ES strict successes, \(\ExcCma\%\) are Excellent; the corresponding shares are \(\ExcQieoReal\%\) for real-coded QIEO and \(\ExcGaBest\%\) for the best GA (Fig.~\ref{fig:stack}).

CMA-ES either locates the basin and converges geometrically, or it does not. QIEO has no equivalent polishing phase; its accuracy is set by the resolution of a thousand observed vectors. The GA concentrates in Fair and Poor and reaches \(10^{-8}\) on \(\bandGEight\%\) of functions.

The strict-success gate of \(10^{-3}\) therefore favours QIEO. The curves cross at \(10^{\flipExp}\); at \(10^{-8}\) CMA-ES leads \(\bandCEight\%\) to \(\bandQEight\%\). Both orderings come from the same runs. A single hit rate at a single gate is a choice of tolerance, not a property of the solvers.

The mechanism is intelligible. CMA-ES, once its covariance has aligned with the local basin, converges geometrically and stops only when the step size underflows, so a run either finds the right basin and polishes it to the last digit or wanders and reports something unrelated. QIEO has no such polishing phase: its accuracy is set by how finely a population of a thousand observed vectors can resolve the neighbourhood of the optimum before the amplitudes saturate. It is a sampler with excellent coverage and a resolution floor, not a converger. The GA, revealingly, is neither, spending most of its mass in the Fair and Poor bands and reaching \(10^{-8}\) on only \(\bandGEight\%\) of names. It only approximates and does not solve.

This has an immediate and uncomfortable consequence for the headline. The strict-success definition is set at roughly \(10^{-3}\), exactly where QIEO has piled up mass and CMA-ES has not. Tightening the criteria for strict-success erodes the advantage that QIEO had. At \(10^{\flipExp}\) the curves cross, and at \(10^{-8}\) CMA-ES leads \(\bandCEight\%\) to \(\bandQEight\%\). 

The best solver of this 256 continuous function suite is therefore not a property of the solvers alone but of the tolerance a practitioner demands, and any comparison that reports a single hit rate at a single gate is reporting a choice as though it were a measurement. We regard this as the most transferable methodological point in the paper. Accuracy profiles should be published alongside success rates, because the two can order the same solvers oppositely.

\subsection{Caveat 3: Specialization}
\subsubsection{Univariate landscape associations}
\label{sec:props}
\begin{figure*}[!t]
\centering
\includegraphics[width=0.95\textwidth]{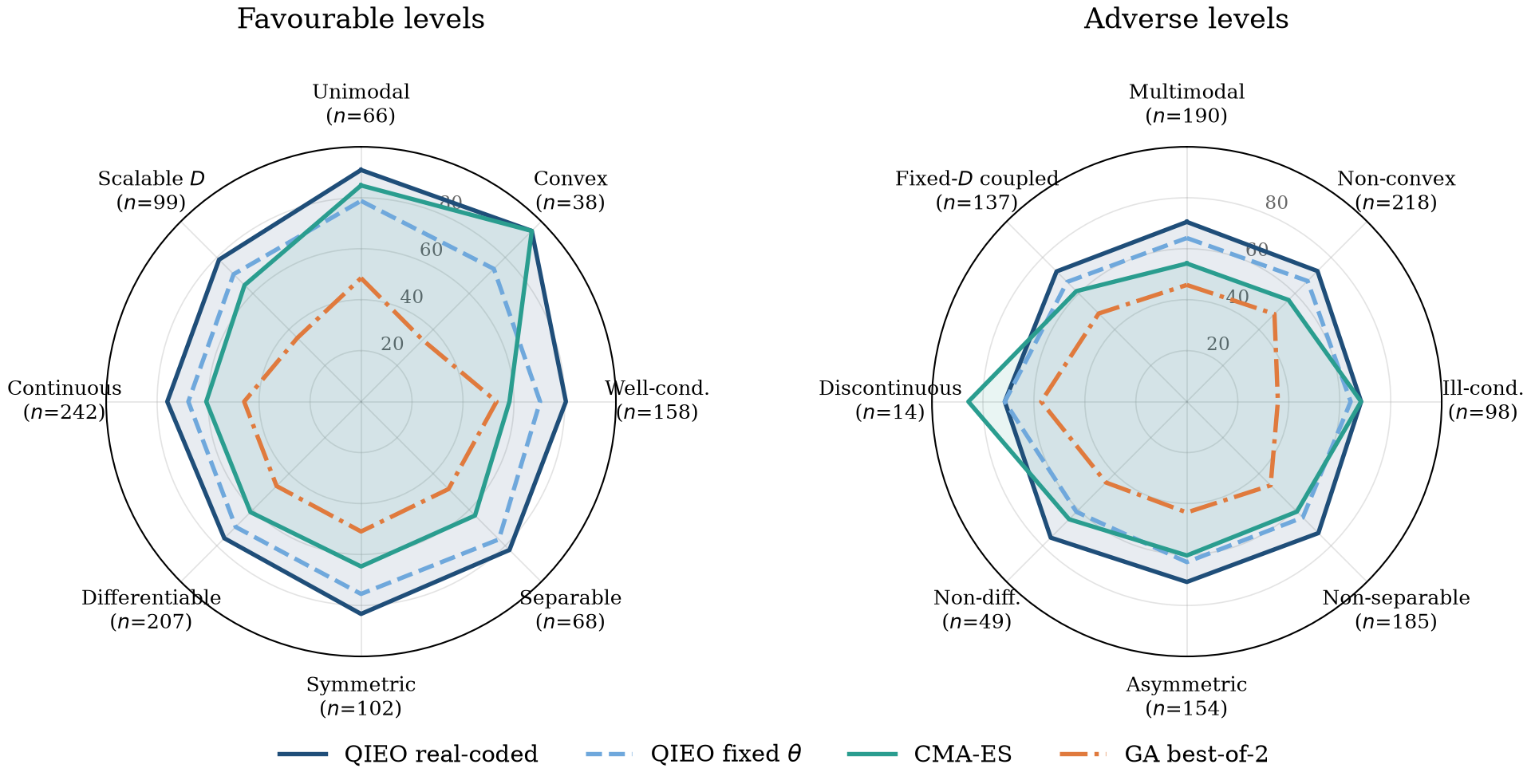}
\caption{Radar profiles of strict success over landscape qualities, favourable levels on the left and adverse levels on the right; each axis is one level and \(n\) is its subset size. Real-coded QIEO and CMA-ES are shaded. The profiles are nested on the favourable levels but change shape on the adverse ones: CMA-ES closes to within two points on ill-conditioning and exceeds QIEO on discontinuity, while the GA retains QIEO's shape at a uniformly lower level.}
\label{fig:radar}
\end{figure*}

Fig.~\ref{fig:radar} shows the strict success rates as a pair of radar profiles, which makes the \emph{shape} of each solver's competence easier to compare. The left panel collects the favourable level of each characteristic and the right panel the adverse level, so a solver that degrades gracefully keeps a similar outline across the two.

On the favourable levels the ordering is uniform and uninteresting: real-coded QIEO encloses every other profile, binary QIEO sits just inside it, CMA-ES inside that except on convexity where the two are exactly tied at \(94.7\%\), and the GA is the smallest polygon on all eight axes. The adverse panel is where the profiles stop being nested and start changing shape, which is the point of plotting them this way. QIEO's margin over CMA-ES is seventeen points on both multimodality and non-convexity, but it narrows to two points on ill-conditioning (\(67.3\%\) against \(65.3\%\)), and on the discontinuous axis CMA-ES pushes outside QIEO's profile altogether (\(85.7\%\) against \(71.4\%\), on only fourteen names). Ill-conditioning is thus the one adverse quality on which the two solvers are genuinely equivalent, and it is the same quality that Section~\ref{sec:portfolio} shows to be the source of CMA-ES's exclusive wins---a hint that the aggregate rate is averaging over two different competences, which Section~\ref{sec:multi} then confirms.

The GA's profile is worth a second look because it is not merely a shrunken version of the others. Its outline follows QIEO's fairly closely in shape while sitting fourteen to thirty-three points inside it, which is what one would expect of two methods that share a coordinate-wise, elite-driven update and differ mainly in how efficiently they exploit it. CMA-ES, by contrast, has a visibly different shape, bulging where QIEO is weak and retreating where QIEO is strong.

Table~\ref{tab:traits2} gives the univariate tests with effect sizes and both raw and Holm-adjusted \(p\)-values. Read naively, five properties look influential. Read with the correction that nine simultaneous tests require, only two survive: multimodality, where QIEO falls from \(90.9\%\) on unimodal names to \(70.0\%\) on multimodal ones (odds ratio \(4.01\), \(p_{\text{Holm}}=0.011\)), and convexity, where it falls from \(94.7\%\) to \(72.0\%\) (odds ratio \(5.70\), \(p_{\text{Holm}}=0.041\)). Conditioning and symmetry, both nominally significant at \(p<0.05\), do not survive (\(p_{\text{Holm}}=0.17\) for each).

\begin{table}[!t]
\caption{Univariate association with real-coded QIEO strict success. OR is the Haldane-corrected odds ratio of the first level against the second. \(p_{H}\) is Holm-adjusted over the nine tests.}
\label{tab:traits2}
\centering
\small
\begin{tabular}{lccrl}
\toprule
Contrast & \% vs \% & OR \([95\%]\) & \(p\) & \(p_{H}\) \\
\midrule
Uni- vs multimodal & 90.9 / 70.0 & 4.01 [1.7,9.5] & 0.0012 & \textbf{0.011} \\
Convex vs not & 94.7 / 72.0 & 5.70 [1.5,21] & 0.0052 & \textbf{0.041} \\
Well- vs ill-cond. & 80.4 / 67.3 & 1.98 [1.1,3.5] & 0.028 & 0.17 \\
Symm. vs asymm. & 83.3 / 70.1 & 2.09 [1.1,3.9] & 0.024 & 0.17 \\
Separable vs not & 82.4 / 72.4 & 1.73 [0.9,3.5] & 0.15 & 0.73 \\
Scalable vs fixed-\(D\) & 78.8 / 73.2 & 1.34 [0.7,2.4] & 0.39 & 1.00 \\
Differentiable vs not & 75.4 / 75.5 & 1.01 [0.5,2.1] & 0.87 & 1.00 \\
Continuous vs not & 75.6 / 71.4 & 1.32 [0.4,4.1] & 0.97 & 1.00 \\
\(f^\star\) depends on \(D\) & 78.6 / 75.2 & 1.09 [0.3,3.7] & 0.97 & 1.00 \\
\bottomrule
\end{tabular}
\end{table}

The negative results are as informative as the positive ones and deserve more attention than they usually get. Differentiability makes no difference whatsoever: \(75.4\%\) on differentiable names against \(75.5\%\) on non-differentiable ones, an odds ratio of \(1.01\). Continuity makes none either. This is exactly what one should expect of a gradient-free, rank-based update, and it falsifies a criticism sometimes levelled at quantum-inspired methods, that they need smooth landscapes to work. They do not. Whatever limits QIEO, it is not the absence of derivatives; it is the arrangement of basins and the size of the space.

\subsubsection{Joint logistic model}
\label{sec:multi}
The labels in Table~\ref{tab:traits} are far from independent. Multimodal functions in this catalogue are also overwhelmingly non-convex, frequently non-separable, and often asymmetric, so a univariate test on any one of them partly measures the others. Table~\ref{tab:logit} fits all of them jointly, together with \(\log_2 D\), for each of the three solver families.

Two things emerge. The first is that the eleven landscape characteristics collapse to roughly two effective axes. Once multimodality and dimension are in the model, conditioning, symmetry, separability and non-differentiability all lose significance for every solver. The catalogue's taxonomy is a useful vocabulary for describing functions, but as a predictive basis it is highly redundant, and stratified tables built on it will keep rediscovering modality and dimension under other names. This is worth saying plainly because such tables are a standard device in benchmark papers.

The second is a clean dissociation between the two leading solvers, and it is the mechanistic heart of the paper. For QIEO, each doubling of the dimension multiplies the odds of success by \(0.33\) (\(p=6\times10^{-5}\)), while ill-conditioning is not significant. For CMA-ES, dimension is almost exactly neutral---an odds ratio of \(1.01\) with \(p=0.96\)---while multimodality is the single dominant term. The GA is the worst case of the QIEO pattern, losing a factor of \(0.18\) in odds per doubling of \(D\).

\begin{table}[!t]
\caption{Joint logistic regression of strict success on labels and \(\log_2 D\) (\(n=256\)). Entries are odds ratios; \(^{**}\) marks \(p<0.01\), \(^{*}\) \(p<0.05\).}
\label{tab:logit}
\centering
\small
\begin{tabular}{lccc}
\toprule
Predictor & QIEO real & CMA-ES & GA best-of-2 \\
\midrule
Multimodal & \(0.24^{**}\) & \(0.21^{**}\) & 0.83 \\
Non-separable & 0.65 & 0.86 & 0.94 \\
Ill-conditioned & 0.70 & 1.63 & 0.54 \\
Asymmetric & 0.72 & 0.76 & 0.91 \\
Non-differentiable & 0.57 & 1.27 & 0.59 \\
\(\log_2 D\) (per doubling) & \(0.33^{**}\) & \(1.01\) & \(0.18^{**}\) \\
\bottomrule
\end{tabular}
\end{table}

\begin{figure}[!t]
\centering
\includegraphics[width=\columnwidth]{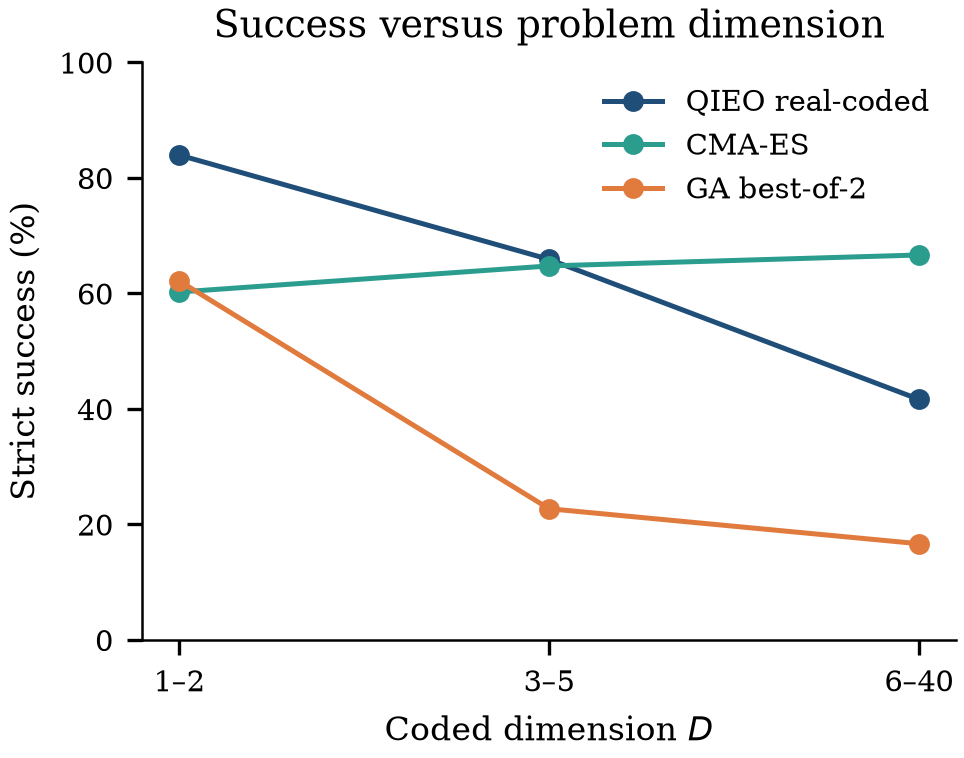}
\caption{Strict success against coded dimension. QIEO's advantage is a low-dimensional phenomenon and disappears by \(D=6\); CMA-ES is nearly flat, and the GA collapses fastest. The highest bin holds only twelve names and should be read as suggestive.}
\label{fig:dim}
\end{figure}

The interpretation follows directly from what the operators can represent. QIEO's advantage is coverage: a thousand simultaneously observed vectors blanket a two-dimensional box densely enough that even a pathologically corrugated landscape is sampled near its global basin. Coverage, however, is exactly the quantity that decays exponentially in \(D\), which is why dimension is QIEO's dominant predictor and why the decay appears even across the modest range this suite offers. CMA-ES makes no attempt at coverage; with \(\lambda=4+\lfloor 3\ln D\rfloor\) it samples a handful of points and invests everything in learning a metric, which is a dimension-robust strategy and precisely why conditioning does not hurt it. What it cannot do is decide which of many basins to learn the metric \emph{of}, so multimodality is its binding constraint. Fig.~\ref{fig:dim} shows the resulting crossover: QIEO leads by twenty-four points at \(D\le 2\), the two draw level at \(D=3\)--\(5\), and CMA-ES leads on the twelve names with \(D\ge 6\).

The honest caveat is that this suite is a poor instrument for dimension: \(156\) names have \(D\le 2\) and only \(12\) have \(D\ge 6\), so the estimate is driven by the contrast between two and five dimensions and the top bin carries wide intervals. The direction is consistent across all three solvers and matches the mechanism, but the magnitude beyond \(D=6\) is an extrapolation, and we would not defend it without a scalable-dimension study.

\subsubsection{Coupling and conditioning}
\label{sec:couple}

\begin{figure*}[!t]
\centering
\includegraphics[width=0.98\textwidth]{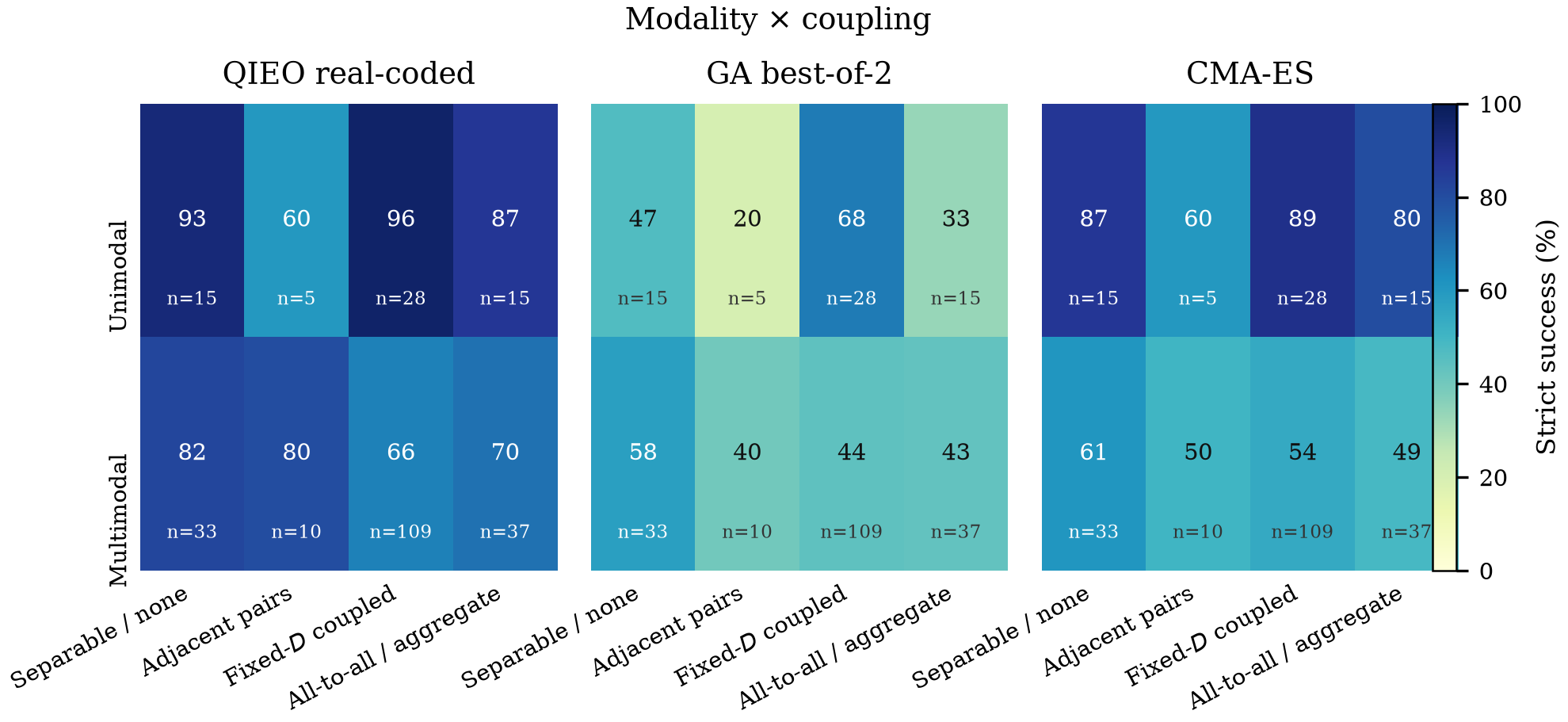}
\caption{Modality against coupling; annotations give subset sizes. QIEO stays competent on separable unimodal names and degrades where basins and cross-coordinate terms coincide. CMA-ES follows the same trend, the GA is uniformly weak.}
\label{fig:couple}
\end{figure*}

\begin{figure*}[!t]
\centering
\includegraphics[width=\textwidth]{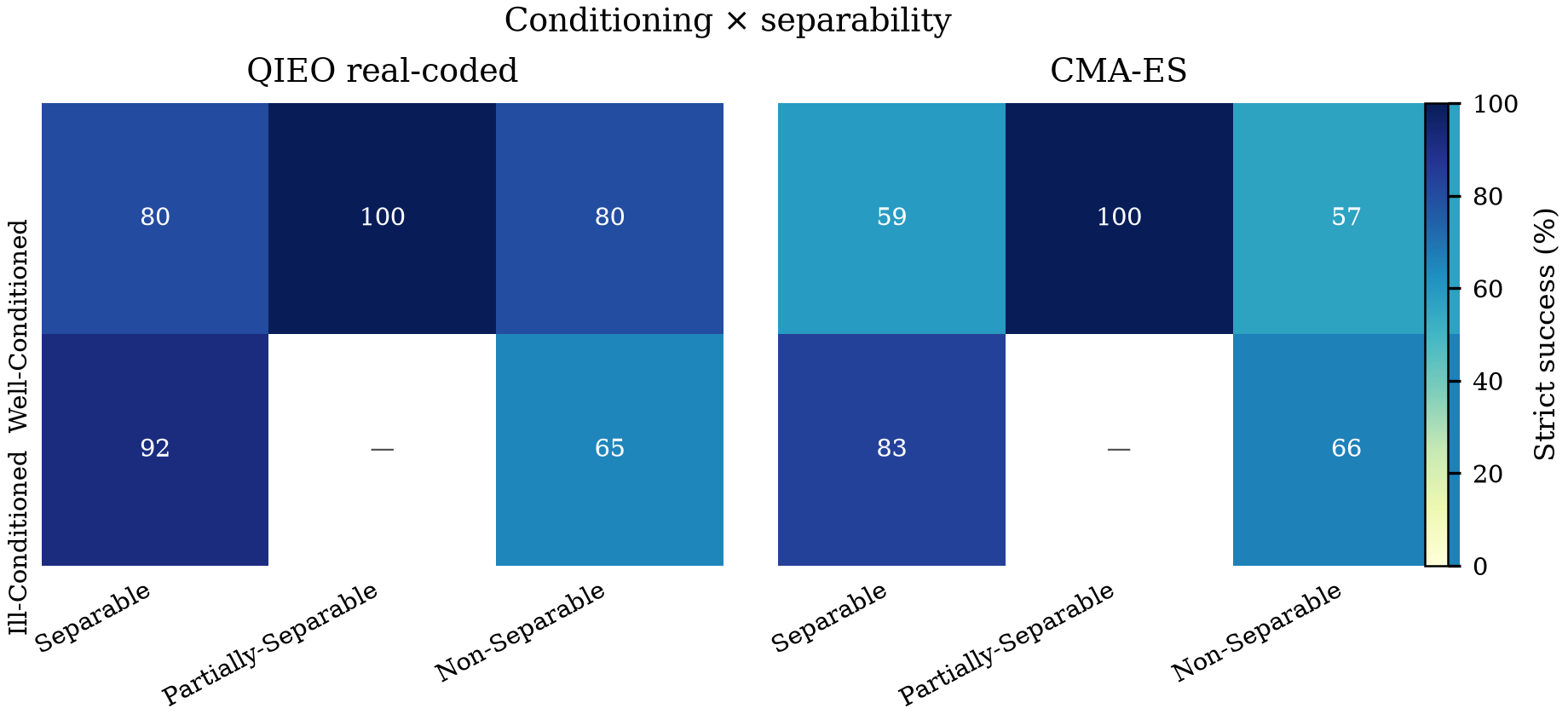}
\caption{Conditioning against separability for real-coded QIEO and CMA-ES. Ill-conditioned non-separable names reduce both solvers without inverting their order.}
\label{fig:condsep}
\end{figure*}

Fig.~\ref{fig:couple} crosses modality with the coupling taxonomy, and Fig.~\ref{fig:condsep} crosses conditioning with separability. Real-coded QIEO is strongest where coordinates are independent and there is a single basin, and weakest where multimodality coincides with all-to-all or fixed-\(D\) coupling. CMA-ES traces the same qualitative pattern at a lower level.

One cell is worth dwelling on because it contradicts an expectation. Adjacent-pair coupling---the Rosenbrock-like sliding residual---is the geometry CMA-ES exists to exploit, and it is the geometry a coordinate-wise rotator should find hardest. On this suite the result runs the other way: QIEO reaches \(73.3\%\) on those fifteen names and CMA-ES only \(53.3\%\). The explanation is population size rather than operator design. A two-dimensional banana valley is a small target, and a default \(\lambda\) of six samples it too sparsely to lock on reliably, whereas a thousand samples find the valley by brute force even without ever representing its direction. Curvature adaptation is the better idea; it is not the better idea at \(\lambda=6\) in two dimensions. This is a concrete instance of the general point that a shared evaluation cap poses different questions to different population sizes.

The conjunction that ought to be QIEO's worst case---non-separable, ill-conditioned \emph{and} multimodal, \(n=\HardQieoN\)---gives \(\HardQieoPct\%\) for QIEO and \(\HardCmaPct\%\) for CMA-ES, with completely overlapping Wilson intervals, while the GA drops to \(31.7\%\). Where the catalogue is hard, it is hard for everything we ran. That is a useful negative result: the difficulty in this region is a property of the objectives rather than a QIEO-specific pathology, and a reader looking for a decisive indictment of the rotation gate will not find it here.

\subsection{Performance comparison of QIEO variants}

\begin{figure*}[!t]
\centering
\includegraphics[width=0.98\textwidth]{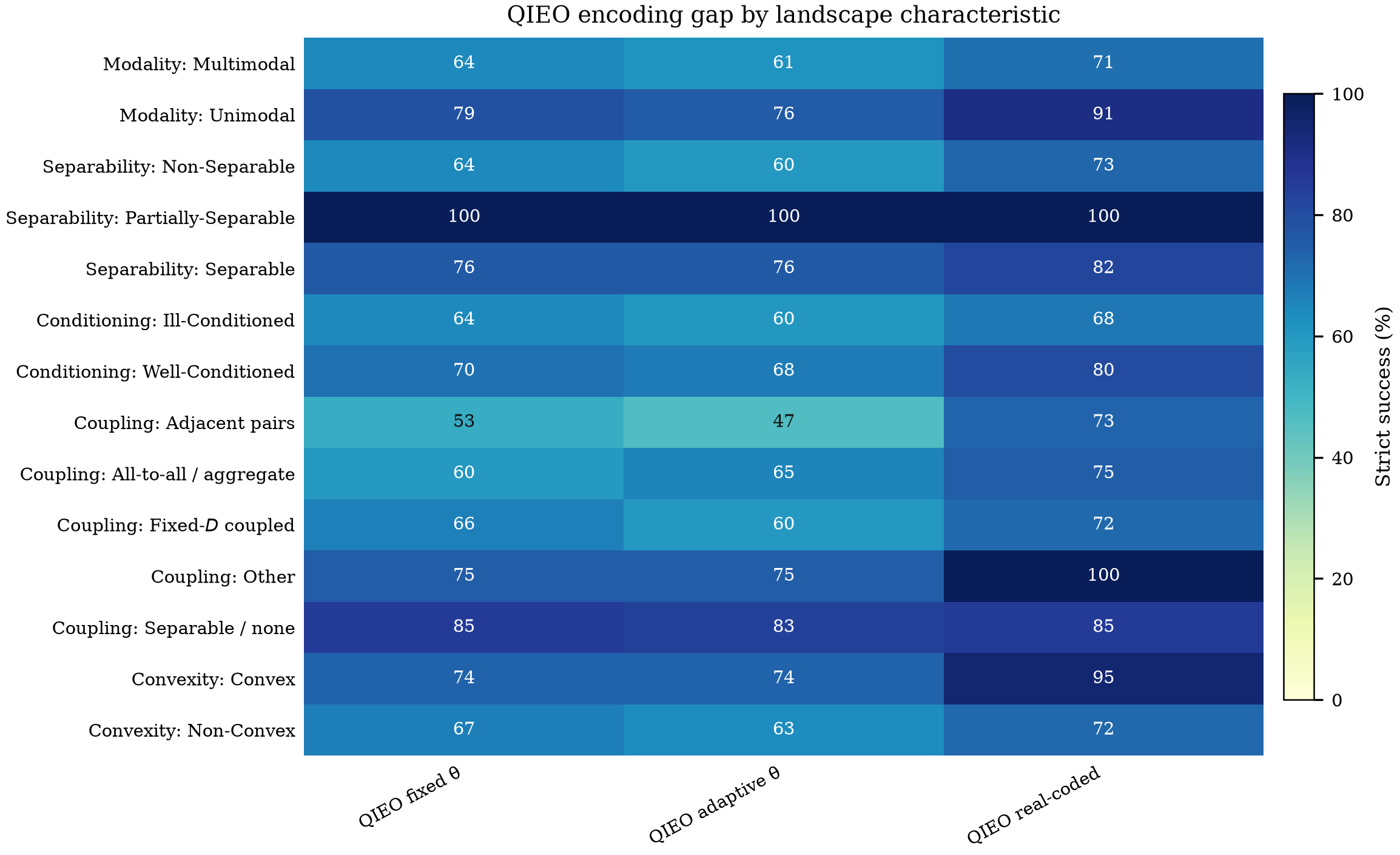}
\caption{QIEO encoding gap across characteristics. Real-coded columns are consistently darker; adaptive \(\theta\) is not a substitute for a real representation.}
\label{fig:enc}
\end{figure*}

Binary QIEO is, structurally, a genetic algorithm wearing a rotation gate, and Fig.~\ref{fig:enc} shows the gate does not cancel the cost of the encoding. Across modality, separability, conditioning, coupling and convexity, the real-coded columns are uniformly darker, and adaptive \(\theta\) never systematically improves on a fixed \(\theta\). What is striking is the comparison in the other direction: even \emph{binary} QIEO, at \(\QieoFixPct\%\), remains far ahead of the binary GA at \(\GaBinPct\%\) with the same bit budget and the same sampling allowance. The rotation-plus-elite update is doing real work that crossover and mutation are not.

The accuracy profiles locate the deficit precisely, and it is not where one might guess. Binary and real QIEO reach machine precision on the same fraction of names, \(\bandQTwelve\%\) each, so the binary encoding is not failing to find basins. It is failing to resolve them: at \(10^{-8}\), real-coded QIEO reaches \(\bandQEight\%\) and the binary variant only \(21.9\%\), and its median residual is nearly three orders of magnitude worse. That is the signature of a quantization floor rather than a search failure. Inspecting the returned solutions confirms it. For \(\gridOn\) of \(\gridChecked\) names, every coordinate of the fixed-\(\theta\) solution lies within two per cent of a cell on the lattice of \(2^{17}\) points spanning the box, and the residual error is of the order of one cell width. Ackley~1 is representative: the box is \([-35,35]\), one lattice step is \(70/2^{17}=5.34\times10^{-4}\), and the reported solution is \(-5.34\times10^{-4}\) in every coordinate, one step from the true minimizer at the origin. The optimizer has done everything its representation permits.

The practical reading is that the binary/real gap in QIEO should not be attributed to the rotation dynamics at all. It is an addressing-resolution limit, and the appropriate remedies are a finer encoding, a local refinement stage, or the real-coded variant, which is what the data already recommend.

\subsection{QIEO versus industry-standard CMA-ES}
\label{sec:portfolio}

\begin{figure}[!t]
\centering
\includegraphics[width=\columnwidth]{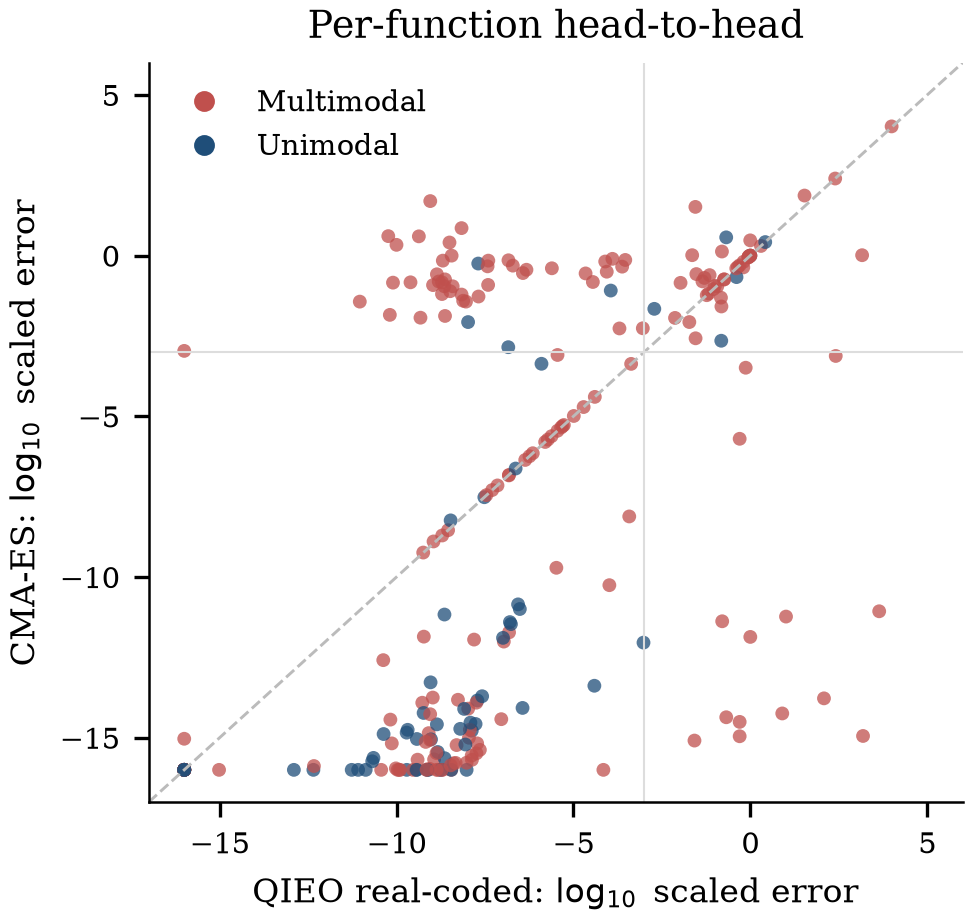}
\caption{Per-function head-to-head, both axes \(\log_{10}\) scaled error, coloured by modality. Points on the lower-left diagonal are solved by both; the horizontal and vertical arms are the exclusive wins that make a two-solver portfolio worthwhile.}
\label{fig:h2h}
\end{figure}

Fig.~\ref{fig:h2h} plots the per-function residual of real-coded QIEO against that of CMA-ES. The bulk of the suite lies in the lower-left corner where both solve to high precision, and the arms along the axes are the interesting part.

Both solvers succeed on \(\nBothQC\) names. QIEO alone succeeds on \(\nOnlyQieo\); CMA-ES alone on \(\nOnlyCma\). Neither list is a random sample of the suite, and their composition is the clearest statement of mechanism in the paper. Thirty-nine of QIEO's \(\nOnlyQieo\) exclusive wins are two-dimensional, and feature densely oscillatory landscapes, namely, Eggholder, Drop-Wave, Shubert~3 and~4, Schaffer~3 and~F6, Odd Square, Easom, De~Jong~5, Ripple, Deflected Corrugated Spring. These are functions with an enormous number of local minima and a global basin that is not distinguished by any large-scale trend, so no metric worth learning exists and the only reliable strategy is to look everywhere. A thousand-member population does exactly that.

Every one of CMA-ES's \(\nOnlyCma\) exclusive wins is labelled multimodal, and eight of the twelve are ill-conditioned. They are Thurber, Ratkowsky~01, De~Villiers--Glasser~1, Biggs EXP6, Chen~V, BBQS, Cosine Mixture, Mishra~1 and~2, Shekel~5, Tripod, and Trid-10. The regression-fit members of that list share a specific geometry, namely, a sum of squared residuals whose Hessian has a condition number in the thousands, producing a long curved trough that is nearly flat along its floor. Coverage is useless there, because the trough is a measure-zero object that random sampling never lands in. The anisotropic step-size adaptation that CMA-ES performs is exactly what is needed and precisely what a coordinate-wise rotation cannot express. Trid-10, an ill-conditioned quadratic at \(D=10\), is the pure form of the same story.

\begin{figure*}[!t]
\centering
\subfloat[Drop-Wave ($D=2$)]{\includegraphics[width=0.48\textwidth]{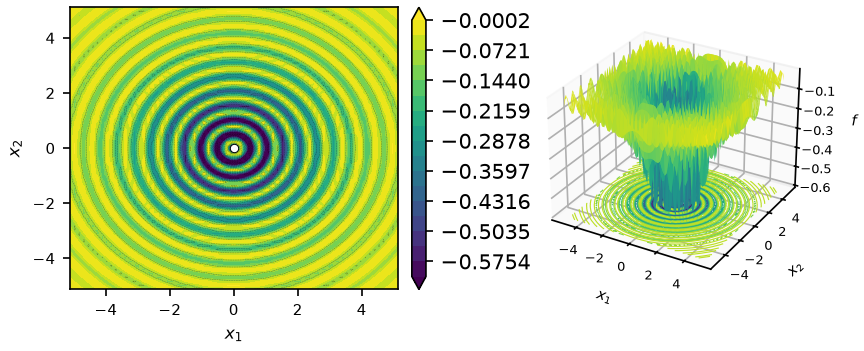}\label{fig:land-drop}}
\hfill
\subfloat[Shubert~3 ($D=2$)]{\includegraphics[width=0.48\textwidth]{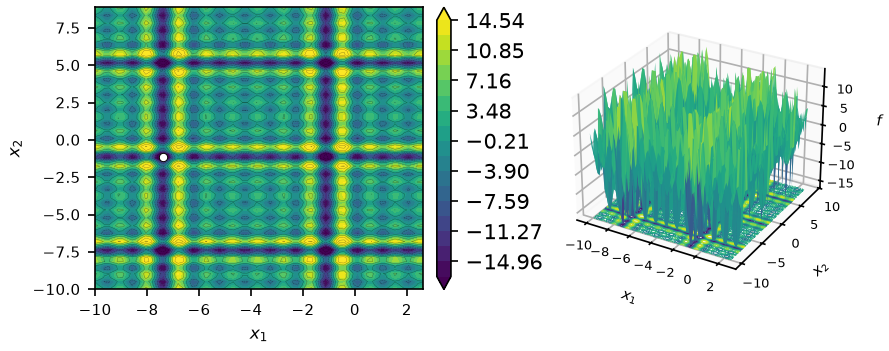}\label{fig:land-shu}}
\caption{Exclusive real-coded QIEO wins (appendix 2-D contour, left of each panel; 3-D surface, right; white marker is campaign-best). Drop-Wave: QIEO real Excellent ($\hat f=-1$), GA best Good, CMA-ES Fair. Shubert~3: QIEO real Good ($\hat f=-29.676$), GA best Good, CMA-ES Poor ($\hat f=-20.29$).}
\label{fig:landq}
\end{figure*}

\begin{figure*}[!t]
\centering
\subfloat[Chen~V ($D=2$)]{\includegraphics[width=0.48\textwidth]{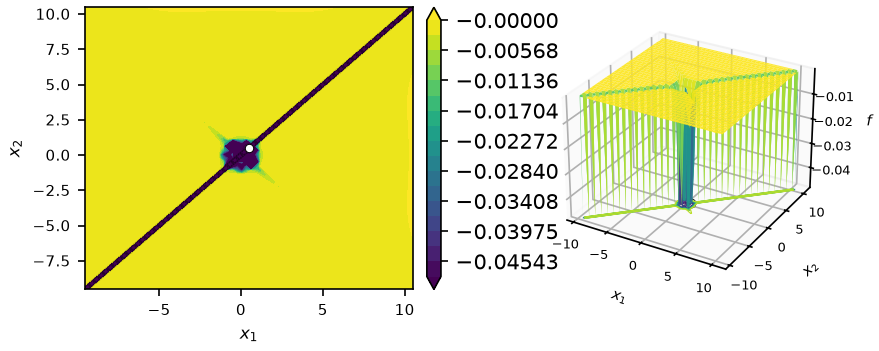}\label{fig:land-chen}}
\hfill
\subfloat[Thurber ($D=7$ slice)]{\includegraphics[width=0.48\textwidth]{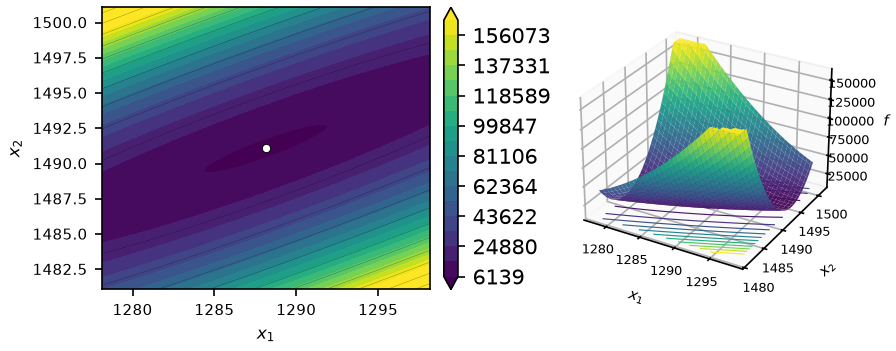}\label{fig:land-thu}}
\caption{Exclusive CMA-ES wins (same appendix layout as Fig.~\ref{fig:landq}). Chen~V: CMA-ES Good ($\hat f=-2000$); every QIEO encoding and both GAs Poor (real-coded QIEO $\hat f=-1000$). Thurber: CMA-ES Excellent ($\hat f=5642.71$); QIEO real Poor ($6.33\times 10^{4}$), GA best Poor.}
\label{fig:landc}
\end{figure*}

Figs.~\ref{fig:landq} and~\ref{fig:landc} make that split visible, using the same 2-D contour and 3-D surface plots as the appendix.

The two competences are therefore close to orthogonal, which is why combining them pays. A portfolio that runs both and keeps the better result reaches \(\nVbsQC\) of 256 names, \(80.1\%\), against \(\QieoRealK\) for the better single solver, and adding both GA encodings lifts this only to \(\nVbsAll\). Given that CMA-ES costs \(\cmaMedEv\) evaluations, running it alongside QIEO is essentially free on the QIEO budget, and a practitioner minimising risk should simply do so. This is a more useful recommendation than any ranking of the two.

\begin{figure}[!t]
\centering
\includegraphics[width=\columnwidth]{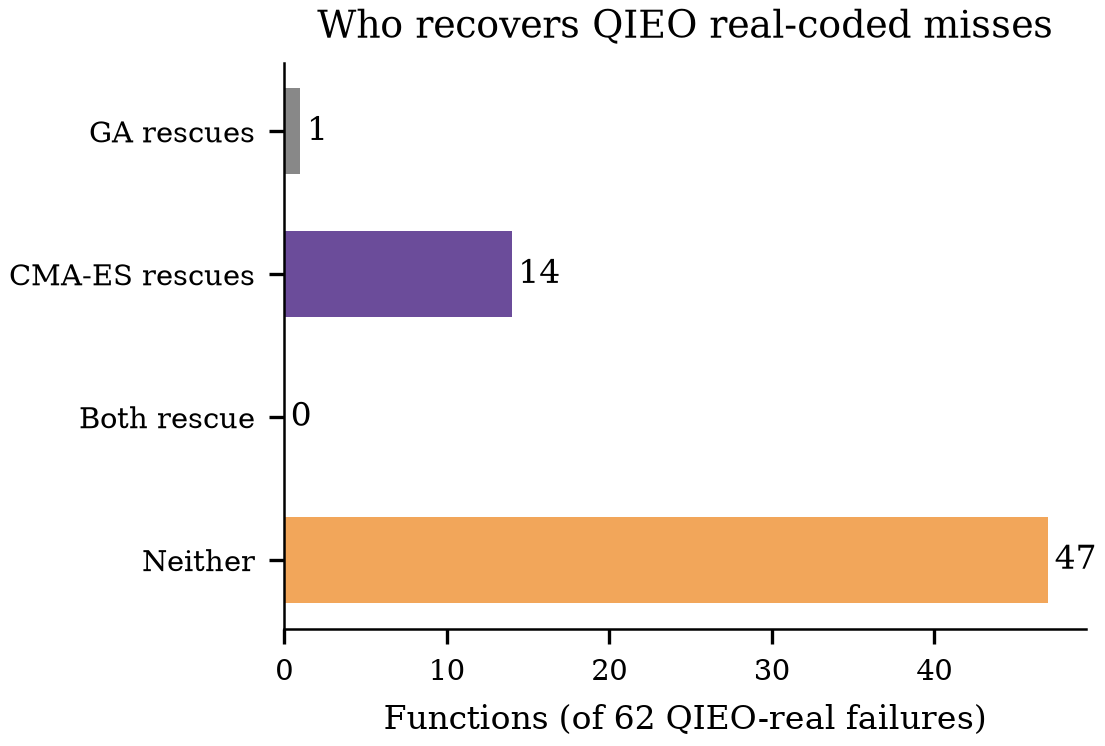}
\caption{Fate of the \(\nQfail\) real-coded QIEO failures. CMA-ES is a specialist rescue on ill-conditioned residuals; the GA recovers one name. Most failures are suite-hard rather than QIEO-specific.}
\label{fig:rescue}
\end{figure}

Fig.~\ref{fig:rescue} shows the same fact from QIEO's perspective. Of the \(\nQfail\) names where real-coded QIEO fails, CMA-ES recovers \(\nCmaRescue\) and the GA recovers exactly \(\nGaRescue\). The ancestor is not a safety net in any meaningful sense; a single recovery out of sixty-three is indistinguishable from noise.

\subsection{Universally unsolved functions}
\label{sec:core}

\(\nCore\) of the 256 objectives are missed by every configuration we ran, while \(\nAllSolved\) are solved by all of them. The unsolved set is the most informative object in the campaign, and its striking feature is what it is \emph{not}: it is not the high-dimensional tail. Its median dimension is \(\coreMedD\), identical to the suite's, and \(47.7\%\) of its members are two-dimensional.

\begin{figure*}[!t]
\centering
\subfloat[Elattar ($D=2$)]{\includegraphics[width=0.48\textwidth]{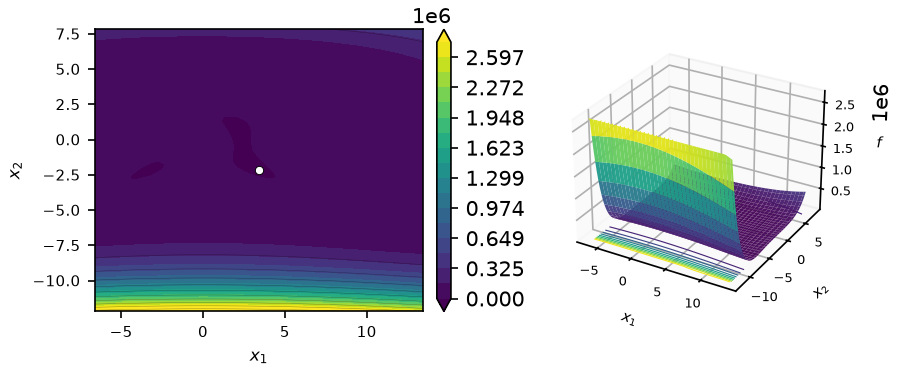}\label{fig:core-ela}}
\hfill
\subfloat[Biggs EXP2 ($D=2$)]{\includegraphics[width=0.48\textwidth]{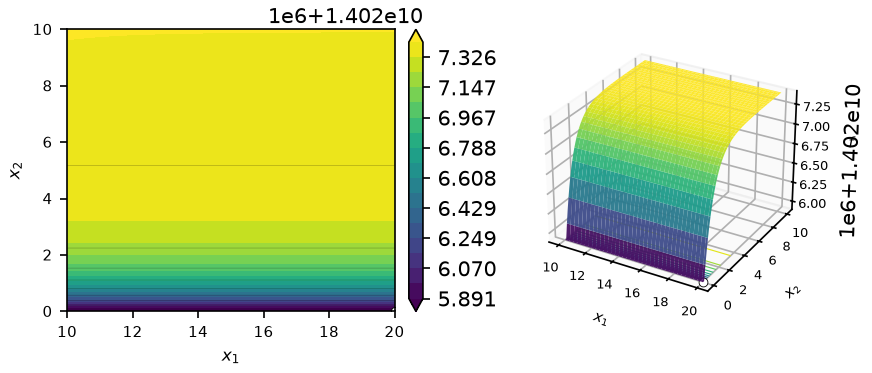}\label{fig:core-b2}}
\caption{Unsolved class~1: nonlinear least-squares residuals (appendix 2-D/3-D). Elattar: all six configurations Poor (QIEO real and CMA-ES both \(\hat f=1.713\)). Biggs EXP2: all six Poor (\(\hat f\sim 1.40\times 10^{10}\)).}
\label{fig:core-nls}
\end{figure*}

\begin{figure*}[!t]
\centering
\subfloat[Cross-Leg-Table ($D=2$)]{\includegraphics[width=0.48\textwidth]{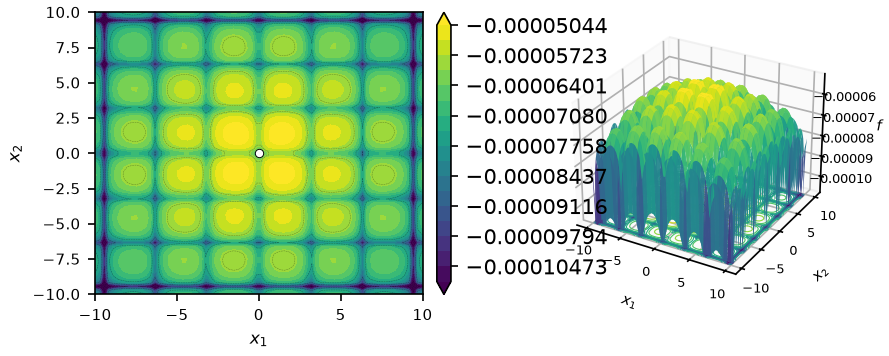}\label{fig:core-clt}}
\hfill
\subfloat[Crowned-Cross ($D=2$)]{\includegraphics[width=0.48\textwidth]{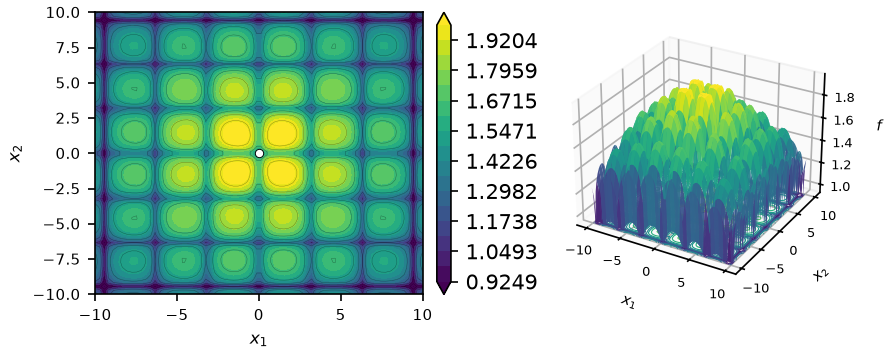}\label{fig:core-cc}}
\caption{Unsolved class~2: the table family (appendix 2-D/3-D). Cross-Leg-Table: all six Poor (CMA-ES best, \(\hat f=-6.71\times 10^{-3}\)). Crowned-Cross: QIEO real Fair (\(\hat f=0.0285\)), CMA-ES Fair (\(\hat f=0.00278\)), both GAs Poor; none reach Good.}
\label{fig:core-table}
\end{figure*}

\begin{figure*}[!t]
\centering
\subfloat[Damavandi ($D=2$)]{\includegraphics[width=0.48\textwidth]{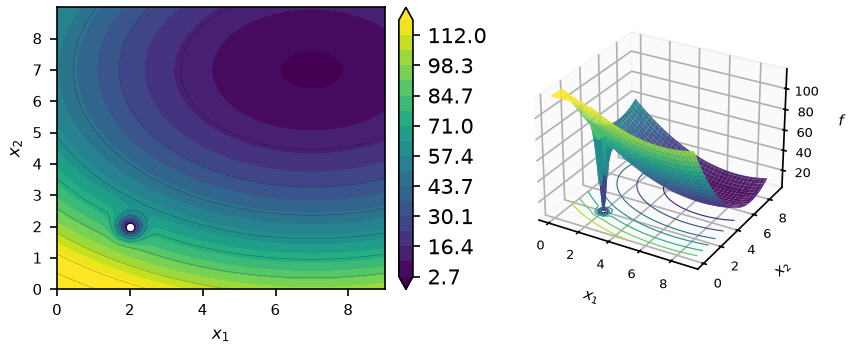}\label{fig:core-dam}}
\hfill
\subfloat[Shubert ($D=2$)]{\includegraphics[width=0.48\textwidth]{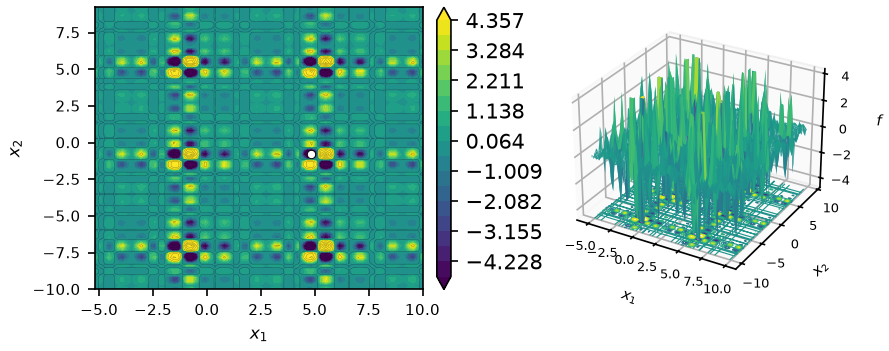}\label{fig:core-shu}}
\caption{Unsolved class~3: classical hard functions (appendix 2-D/3-D). Damavandi: QIEO real Poor (\(\hat f=2\)), CMA-ES Poor (\(\hat f=2\)), GA binary Fair. Shubert: all six Poor (QIEO real \(\hat f=-18.10\) against \(f^\star=-186.73\)).}
\label{fig:core-hard}
\end{figure*}

What the set has instead is a family signature. Ten of the core are nonlinear least-squares data-fitting problems---Biggs EXP2 through EXP5, Brad, Hougen, Watson, De~Villiers--Glasser~2, Elattar, Multi-Gaussian---in which the objective is a sum of squared residuals against tabulated measurements, the parameters differ by many orders of magnitude, and the trough is both curved and nearly flat (Fig.~\ref{fig:core-nls}). Meyer, Meyer--Roth and Osborne are Poor on every solver, but they do not belong in that family count: the published \(x^\star\) lies outside the coded box, so the failure is an infeasible reference rather than an unsolved trough (they are the same three names omitted from the shift campaign). A second cluster is the ``table'' family of flat plateaus punctuated by narrow spikes: Cross-Leg-Table, Crowned-Cross, Holder-Table~2, Table~1 and~2 (Fig.~\ref{fig:core-table}). A third is the classical hard set: Rosenbrock (coded at \(D=4\)), Griewank, Shubert, Rana, Whitley, Damavandi, Bukin~6, Langerman~5, Xin-She~Yang~2 and~4 (Fig.~\ref{fig:core-hard}). Rosenbrock belongs here as run: every configuration is Poor, the best residual being \(0.208\) on real-coded QIEO against \(f^\star=0\). Consistent with this, the only enrichments that reach nominal significance are non-convexity (\(97.7\%\) of the core against \(85.2\%\) of the suite, \(p=0.019\)) and fixed-\(D\) coupling (\(68.2\%\) against \(53.5\%\), \(p=0.048\)); multi-modality and ill-conditioning are elevated but not significantly so.

The implication for benchmark design is worth stating. Difficulty in this catalogue is not a scalar that grows with dimension or modality; it is concentrated in identifiable problem families whose geometry defeats coverage and metric learning simultaneously. A suite that reports mean performance over 256 names is largely reporting how many members of these families it happens to contain, and two suites with different family mixes will rank the same solvers differently for reasons that have nothing to do with the solvers. Reporting per-family results, as we do in the Appendix, is the cheapest available remedy.

\subsection{Shifted comparison}
A reasonable objection to any result on the continuous functions considered is that their minimizers sit at convenient places, such as, the origin, the box centre, or a corner. This may favour coordinate-wise methods, providing an unearned advantage from sampling those points early. The concern is not hypothetical as demonstrated by \(18\) of QIEO's returned solutions are at the origin exactly. The Gavana shift tests it directly by relocating each tabulated \(x^\star\) to a new interior point while leaving \(f^\star\) unchanged.

\begin{figure*}[!t]
\centering
\includegraphics[width=\columnwidth]{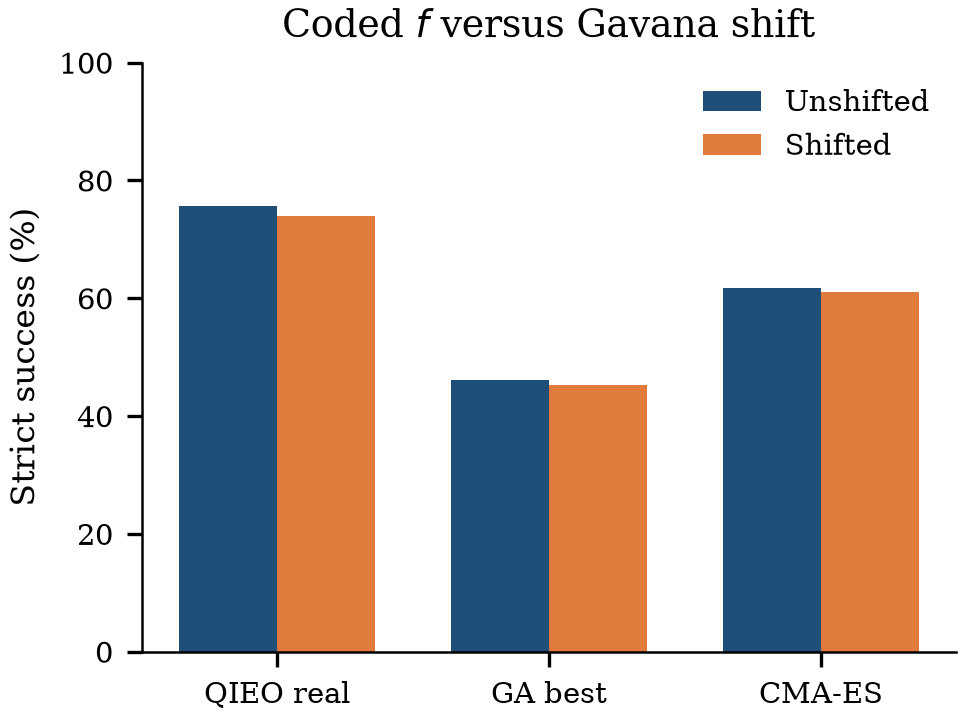}
\caption{Strict success on the 251 names admitting a tabulated \(x^\star\), coded against Gavana-shifted. The ranking and the rates are both essentially unchanged.}
\label{fig:shift}
\end{figure*}

Superficially, the effect is essentially nil (Fig.~\ref{fig:shift}). Real-coded QIEO moves from \(75.4\%\) to \(74.5\%\), the GA from \(46.1\%\) to \(45.4\%\), and CMA-ES from \(60.9\%\) to \(61.0\%\). However, at a finer level, an interesting behaviour is observed. Based on the defined rules for strict-success, QIEO loses thirteen functions and gains nine, CMA-ES loses eighteen and gains seventeen.

The flip from strict success to failure and vice versa is balanced, i.e. wins and losses are nearly equal. If QIEO’s lead came from sitting on convenient points (origin, centre, corner), shifting those points inside the box should produce a one-way drop resulting in many losses and few gains. Balanced turnover indicates that the shift mostly reshuffles which basins get found, not that it strips a systematic advantage.

We take this as reasonably strong evidence against the origin-bias explanation for QIEO's lead, and it is worth reporting precisely because it is a negative result that a sceptical reader would otherwise have to assume.

\subsection{Suspect published minima}

The Bad label is assigned when \(\hat f\) undercuts \(f^\star\) by more than the Good slack. Agreement across independent solvers implicates the published reference. 
\begin{itemize}
\item Holder-Table~2 and Multi-Gaussian report "Bad" by all six configurations
\item Cosine Mixture, Mishra~1, and Mishra~2 report "Bad" by five out of six configurations
\item Rana report "Bad" by four out of six configurations
\item Watson reports "Bad" by CMA-ES alone and is treated as a solver artefact. 
\end{itemize}
These six remain in the rates above. Three of them are Gavana additions.

\section{Discussion}
\label{sec:disc}
The first observation presented in Sec.\ref{Sec:success_rate} is extremely misleading. Fig.\ref{fig:overall}, which presents strict success rate, suggests that real-coded QIEO (\(\QieoRealPct\%\)) outperforms both CMA-ES (\(\CmaPct\%\)) and GA (\(\GaBestPct\%\)) . However, there are three main caveats that must be considered. 

Firstly, the arithmetic cost that QIEO needs to accommodate, in terms of number of function evaluations, which has been capped at $5e^5$ (1000 potential candidates per generation, capped at 5000 maximum generations) in this study is vastly greater than that required by CMA-ES.

The first is arithmetic about cost. CMA-ES terminates on its own convergence criteria after a median of \(\cmaMedEv\) function evaluations, which is \(\cmaCapFrac\%\) of the budget QIEO is entitled to spend. QIEO, stopping on fitness stagnation, needs a median of \(\qieoMedEv\) evaluations, so the asymmetry is a property of the search and not of a budget it never used. Normalized per evaluation, CMA-ES solves \(\cmaPerM\) names per million evaluations and QIEO at most \(\qieoPerM\), a gap of two to three orders of magnitude that no amount of aggregate hit rate can hide. 
The second concerns what ``success'' means. The two solvers do not merely differ in how often they succeed; they fail in qualitatively different shapes. CMA-ES is close to bimodal, driving \(\bandCTwelve\%\) of the suite to machine precision or missing the basin outright, whereas QIEO deposits a great deal of mass in the intermediate band that the conventional \(10^{-3}\) gate happens to reward. Slide the gate to \(10^{\flipExp}\) and the ranking reverses. The third is about where each solver actually earns its wins. QIEO's \(\nOnlyQieo\) exclusive victories are dominated by low-dimensional, densely multimodal landscapes, such as, Eggholder, Drop-Wave, Shubert, Schaffer, on which a thousand-member population behaves like a dense grid. Every one of CMA-ES's \(\nOnlyCma\) exclusive victories is multimodal, and most are ill-conditioned, namely, Thurber, Ratkowsky, De Villiers--Glasser, Biggs EXP6, Trid-10. These are two different competences, not two positions on one scale.

Underneath, the labelling itself turns out to be largely redundant. A multivariate model of QIEO success on the binary traits retains only multimodality and dimension; conditioning, symmetry, separability and differentiability lose all explanatory power once those two are present, and only two of nine univariate tests survive correction for multiple comparisons. More striking is a clean dissociation: dimension is the strongest predictor for QIEO and statistically irrelevant for CMA-ES, while multimodality is the only surviving predictor for CMA-ES. The two algorithms have almost orthogonal weaknesses, which is precisely why a portfolio of the two reaches \(\nVbsQC/256\) when the better single solver reaches \(\QieoRealK\).

\paragraph{QIEO vs. CMA-ES}
The single idea that organizes every result above is that QIEO and CMA-ES offer different strengths. QIEO excels in coverage. Its operator is too weak to model a landscape, but a thousand simultaneous observations blanket a low-dimensional box thoroughly, and on a corrugated two-dimensional surface with no exploitable large-scale structure that is the winning strategy, which is why forty of its forty-nine exclusive wins are two-dimensional oscillators. CMA-ES relies on curvature. It samples sparsely and spends its budget learning a metric, which is worthless when there is no metric to learn but decisive on a curved, ill-conditioned trough, which is why its exclusive wins are the regression residuals. 

Coverage decays exponentially in dimension and is indifferent to conditioning; curvature is indifferent to dimension and defeated by multi-modality. Table~\ref{tab:logit} quantitatively validates this statement and establishes that the aggregate ranking becomes the least interesting thing in the study.

\paragraph{Applicability of QIEO}
Real-coded QIEO is a good choice when \(D\) is small, evaluations are cheap and parallel, the landscape is densely multimodal, and a residual near \(10^{-6}\) is acceptable. It is a poor choice when evaluations are expensive, when \(D\) grows, or when the answer must be exact to machine precision. Use the real encoding: the binary variants are limited by an addressing floor of one part in \(2^{17}\) of the box, not by their dynamics, and no amount of \(\theta\) adaptation repairs that. Because CMA-ES costs a median of \(\cmaMedEv\) evaluations against QIEO's millions, there is no scenario in which running both is not worth it; the portfolio reaches \(80.1\%\) where either alone reaches at most \(\QieoRealPct\%\).

\paragraph{What this means for reporting benchmarks.}
Three of our findings are methodological rather than algorithmic, and we think they generalize beyond QIEO. A shared evaluation \emph{cap} does not equalize expenditure when population sizes differ by two orders of magnitude, so cost-normalized rates belong next to raw ones. A single accuracy gate can reverse a ranking: ours reverses at \(10^{\flipExp}\), and reporting only the \(10^{-3}\) rate would have concealed that CMA-ES is the more accurate solver wherever it works at all. And a taxonomy of eleven hand-assigned properties, stratified one at a time, mostly rediscovers two underlying axes; without a joint model it is easy to publish five significant properties where two exist.

\paragraph{Where the GA leaves the picture.}
QIEO was motivated as a more disciplined genetic algorithm, and on this evidence the motivation is fully vindicated in a way we did not anticipate. The GA is not merely lower on hit rate; it is Pareto-dominated, spending a median of \(\gaMedEv\) evaluations, more than a hundred times CMA-ES's, to reach a rate fifteen points lower. Its accuracy profile shows why: it accumulates Good and Fair grades and almost never converges tightly, reaching \(10^{-8}\) on \(\bandGEight\%\) of names against CMA-ES's \(\bandCEight\%\). It approximates rather than solves, it recovers one of sixty-three QIEO failures, and it is weaker in every characteristic slice we plotted. Whatever else is arguable here, the ancestor is not a fallback.

\paragraph{Threats to validity.}
Several are worth naming. The comparison uses one run per (name, encoding), so the intervals are binomial over the suite and not over seeds; a name near a gate boundary could plausibly flip on a different seed, though it is unlikely that many would flip in the same direction. CMA-ES is the textbook default with restarts disabled, which is the honest baseline for ``what a practitioner reaches for'' but understates what the algorithm family can do; IPOP or BIPOP under the same cap would very likely close much of the multimodal gap, and would still cost a small fraction of QIEO's budget. That experiment is the single most valuable follow-up, and until it is run our multimodality result should be read as a statement about default CMA-ES rather than about covariance adaptation. The dimension result rests on twelve names above \(D=5\). The labels are expert tags rather than sampled ELA features~\cite{mersmann2011ela}, so they carry the biases of the sources they came from. And the gates inherit any error in the published minima, which is why we report the six suspect references explicitly rather than quietly dropping them.

\paragraph{What would falsify the account.}
The coverage-versus-curvature reading makes specific predictions, and they are testable. A QIEO variant with a covariance-aware or otherwise rotation-invariant update should recover the ill-conditioned regression names that currently belong to CMA-ES alone, without losing the two-dimensional oscillators. A niching or multi-elite variant should extend the coverage advantage further into moderate dimension. Conversely, a restart CMA-ES should absorb most of QIEO's exclusive multimodal wins while keeping its cost advantage, and if it does, the correct summary of this paper becomes that QIEO buys with population what restarts buy with sequence. Any of these outcomes would sharpen the account; none of them is implied by the aggregate table with which we began.

\section{Conclusion}

QIEO's limitation on complex continuous objectives is real, specific, and not a story about gradients. Non-differentiability and discontinuity cost it nothing at all; multimodality and, above all, dimension cost it a great deal, and the eleven-label taxonomy that seemed to explain its failures collapses to those two axes once they are fitted jointly. Against its ancestor the verdict is unambiguous: the GA is dominated on cost and on quality at once, and rescues one of sixty-three QIEO failures. Against the industrial standard the verdict is not a ranking but a division of labour. Real-coded QIEO records the higher hit rate on this suite, at a gate of \(10^{-3}\) and, even when allowed to stop as soon as it stagnates, at two orders of magnitude more sampling per name; CMA-ES is the more accurate solver wherever it works, overtakes QIEO at any tolerance below \(10^{\flipExp}\), is insensitive to dimension, and uniquely solves the ill-conditioned regression residuals that coverage cannot reach. The two are complementary rather than competing, cheaply combined, and best reported together. Forty-four names in the catalogue resist all of them, clustered by problem family rather than by difficulty, and six carry published minima that our runs contradict. The Appendix reprints the coded objectives and the campaign tables from which every number here is computed.

\section*{Acknowledgment}

\appendices
\section{Property-wise strict-success rates}
\label{app:rates}
Table~\ref{tab:rates} lists the heatmap row values. Intervals are Wilson \(95\%\) on the QIEO real-coded column.

\begin{table*}[!t]
\caption{Strict success (\%) by landscape label. \(n\) is the number of unshifted names in the row.}
\label{tab:rates}
\centering
\small
\begin{tabular}{llrcccc}
\toprule
Property & Level & \(n\) & QIEO real & QIEO CI & GA best & CMA-ES \\
\midrule
Continuity & Continuous & 242 & 75.6 & \([69.8,80.6]\) & 45.5 & 59.5 \\
Continuity & Discontinuous & 14 & 71.4 & \([45.4,88.3]\) & 57.1 & 85.7 \\
Differentiability & Differentiable & 207 & 75.4 & \([69.1,80.7]\) & 46.4 & 59.9 \\
Differentiability & Non-differentiable & 49 & 75.5 & \([61.9,85.4]\) & 44.9 & 65.3 \\
Separability & Separable & 68 & 82.4 & \([71.6,89.6]\) & 48.5 & 63.2 \\
Separability & Partially separable & 3 & 100.0 & \([43.8,100]\) & 0.0 & 100.0 \\
Separability & Non-separable & 185 & 72.4 & \([65.6,78.4]\) & 45.9 & 59.5 \\
Scalability & Scalable & 99 & 78.8 & \([69.7,85.7]\) & 35.4 & 64.6 \\
Scalability & Non-scalable & 157 & 73.2 & \([65.8,79.6]\) & 52.9 & 58.6 \\
Modality & Unimodal & 66 & 90.9 & \([81.6,95.8]\) & 48.5 & 84.8 \\
Modality & Multimodal & 190 & 70.0 & \([63.1,76.1]\) & 45.3 & 52.6 \\
Convexity & Convex & 38 & 94.7 & \([82.7,98.5]\) & 34.2 & 94.7 \\
Convexity & Non-convex & 218 & 72.0 & \([65.7,77.6]\) & 48.2 & 55.0 \\
Conditioning & Well-conditioned & 158 & 80.4 & \([73.5,85.8]\) & 53.2 & 58.2 \\
Conditioning & Ill-conditioned & 98 & 67.3 & \([57.6,75.8]\) & 34.7 & 65.3 \\
Symmetry & Symmetric & 102 & 83.3 & \([74.9,89.3]\) & 51.0 & 64.7 \\
Symmetry & Asymmetric & 154 & 70.1 & \([62.5,76.8]\) & 42.9 & 58.4 \\
Maximum dimensionality & Native 2-D & 120 & 80.0 & \([72.0,86.2]\) & 60.8 & 60.8 \\
Maximum dimensionality & Native 3--5-D & 32 & 50.0 & \([33.6,66.4]\) & 21.9 & 50.0 \\
Maximum dimensionality & Native \(\geq\)6-D & 10 & 30.0 & \([10.8,60.3]\) & 10.0 & 50.0 \\
Maximum dimensionality & Scalable \(D\) & 94 & 83.0 & \([74.1,89.2]\) & 39.4 & 66.0 \\
\(f^\star\) depends on \(D\) & No & 242 & 75.2 & \([69.4,80.2]\) & 45.5 & 60.7 \\
\(f^\star\) depends on \(D\) & Yes & 14 & 78.6 & \([52.4,92.4]\) & 57.1 & 64.3 \\
Variable coupling & Separable / none & 48 & 85.4 & \([72.8,92.8]\) & 54.2 & 68.8 \\
Variable coupling & Adjacent pairs & 15 & 73.3 & \([48.0,89.1]\) & 33.3 & 53.3 \\
Variable coupling & Fixed-\(D\) coupled & 137 & 71.5 & \([63.5,78.4]\) & 48.2 & 59.1 \\
Variable coupling & All-to-all / aggregate & 52 & 75.0 & \([61.8,84.8]\) & 40.4 & 57.7 \\
\bottomrule
\end{tabular}
\end{table*}

\section{Universally unsolved names}
\label{app:core}
The following names are not a strict success for any QIEO encoding, either GA encoding, or CMA-ES: Brad, Biggs EXP2--EXP5, Bohachevsky~2, Bukin~6, Chen Bird, Chichinadze, Cola, Damavandi, De~Villiers--Glasser~2, Elattar, Giunta, Griewank, Hansen, Langerman~5, Mishra~4, Rana, Rosenbrock, Rosenbrock Modified, Shubert, Step-Int, Stretched~V, Table~1, Table~2, Trefethen, Watson, Whitley, Xin-She~Yang~2, Xin-She~Yang~4, Cross-Leg-Table, Crowned-Cross, Happy-Cat, Holder-Table~2, Hougen, Multi-Gaussian, New-Function~2, Picheny, XOR, and Zagros. Meyer, Meyer--Roth and Osborne are likewise Poor on every solver, but the published \(x^\star\) is outside the coded box; they are listed separately from the landscape core in Section~\ref{sec:core}.

\section{Coded catalogue and campaign tables}
\label{app:catalogue}
The following pages reprint the 256 coded objectives---equation, box, dimension, published \(f^\star\), two-dimensional visualization, and unshifted plus shifted campaign tables for every QIEO encoding, both GA encodings, and CMA-ES. Equations follow \texttt{index.py}. Five shifted names without a tabulated \(x^\star\) in the box are marked skipped.

\section{Extra stuff}
Let the binary substring associated with coordinate \(x_j\) be interpreted as the unsigned integer
\begin{equation}
    z_j
    =
    \sum_{r=0}^{B-1} b_{j,r}2^r,
    \qquad
    z_j \in [0,2^B-1].
    \label{eq:qieo_binary_integer}
\end{equation}
The decoded real-valued coordinate is then obtained through fixed-point mapping:
\begin{equation}
    x_j
    =
    l_j
    +
    \frac{z_j}{2^B-1}
    \left(u_j-l_j\right).
    \label{eq:qieo_binary_decoding}
\end{equation}
This mapping ensures that binary QIEO samples the prescribed box-constrained design space with a uniform representational resolution of
\begin{equation}
    \delta_j
    =
    \frac{u_j-l_j}{2^B-1}.
    \label{eq:qieo_resolution}
\end{equation}

\end{document}